\documentclass[lettersize,journal]{IEEEtran}
\usepackage{amsmath,amsfonts}
\usepackage{algorithmic}
\usepackage{algorithm}
\usepackage{array}
\usepackage[caption=false,font=normalsize,labelfont=sf,textfont=sf]{subfig}
\usepackage{textcomp}
\usepackage{stfloats}
\usepackage{url}
\usepackage{verbatim}
\usepackage{graphicx}
\usepackage{cite}

\usepackage{stmaryrd}
\DeclareMathOperator{\vect}{vec}
\DeclareMathOperator{\fold}{fold}
\DeclareMathOperator{\unfold}{unfold}
\DeclareMathOperator{\bcirc}{bcirc}

\begin{document}

\title{Data Understanding Survey: Pursuing Improved Dataset Characterization Via Tensor-based Methods}

\author{Matthew D. Merris, Tim Andersen
\thanks{This work was supported by the National Science Foundation (Award Num: 1849463 \& 2223932), the Laboratory for Physical Sciences, and Boise State University - School of Computing. (Corresponding author: Matthew D. Merris.)}
\thanks{Matthew D. Merris (email:{matthewmerris@u.boisestate.edu}) is with the School of Computing, Boise State University, Boise, Idaho 83702, USA.}
\thanks{Tim Andersen is with the Department of Computer Science, Boise State University, Boise, Idaho 83702, USA.}
}

\markboth{Journal of \LaTeX\ Class Files,~Vol.~14, No.~8, August~2021}%
{Shell \MakeLowercase{\textit{et al.}}: A Sample Article Using IEEEtran.cls for IEEE Journals}


\maketitle

\begin{abstract}
In the evolving domains of Machine Learning and Data Analytics, existing dataset characterization methods such as statistical, structural, and model-based analyses often fail to deliver the deep understanding and insights essential for innovation and explainability. This work surveys the current state-of-the-art conventional data analytic techniques and examines their limitations, and discusses a variety of tensor-based methods and how these may provide a more robust alternative to traditional statistical, structural, and model-based dataset characterization techniques. Through examples, we illustrate how tensor methods unveil nuanced data characteristics, offering enhanced interpretability and actionable intelligence. We advocate for the adoption of tensor-based characterization, promising a leap forward in understanding complex datasets and paving the way for intelligent, explainable data-driven discoveries.
\end{abstract}


\section{Introduction}
\label{sec:intro}
The advent of general-purpose computing initiated a continuous escalation in the complexity and scale of computer systems, algorithms, and data. The ubiquity of computing in nearly all domains of modern life underscores its role as a fundamental driver of innovation and discovery. Computational systems have consistently surpassed performance expectations across diverse application domains, enabling analyses and insights that were previously unattainable. However, the limitations of classical computing architectures are becoming increasingly apparent, prompting system architects to move beyond the traditional von Neumann model in favor of heterogeneous computing paradigms better suited to contemporary computational demands. Notably, advances such as Google’s fourth-generation Tensor Processing Unit (TPU) \cite{jouppiTenLessonsThree2021} have demonstrated the feasibility of exaflop-scale computation in commercial settings, alongside the development of TPU-based designs optimized for edge computing applications \cite{seshadriEvaluationEdgeTPU2022}.

Concomitant with the ever-increasing complexity of hardware and software algorithms, data has become massive in scale and complexity, driven in part by the successes of implicit models in research and industry. 
Implicit models infer parameterization directly from data as opposed to explicit models which are parameterized in advance and manually fine-tuned.
The data driving these types of applications tends to be either structured (e.g. dates, user names, geolocations, etc), unstructured (e.g. bodies of text, audio recordings, EEGs, etc) or some combination of both \cite{dzeroskiMultirelationalDataMining2003}.

Dataset characterization (DC) provides a framework for the systematic analysis of datasets based on a set of measures directly calculated from a dataset.
Historically DC has been used as a means of ensuring dataset quality and relevance in data-centric applications by providing understanding of data integrity, anomaly detection, pattern recognition, and distributions of data.
DC has evolved in the space of modern computing to include data preprocessing for machine learning (ML) models, the design of architectures for big data systems, and to ensure compliance with data governance standards.  
In the context of artificial intelligence (AI), DC is used to evaluate datasets for representativeness, biases, and enhancing the performance of predictive models.
The evolution of DC reflects the increasing complexity and scale of data-driven application in the modern digital age.

\subsection{Tensors, Utilitarian Tools for Modern Computing}
Tensors are a mathematical abstraction that generalize matrices and vectors to an arbitrary order (e.g. number of indices or modes) that embody the concepts of equivariance, multilinearity, and separability \cite{limTensorsComputations2021}. 
As such, they are a versatile tool that has played a role in many of the major scientific and engineering advancements from the last two centuries.  

In terms of modern computing, the tensor is a convenient representation for multi-relational datasets leading to a proliferation of tensor-based methods across a spectrum of problem domains.
These methods constitute a framework for tensor data analysis (TDA). 
The utility of tensors extends beyond themselves alone which is exemplified by their role in several of the so-called 'top ten algorithms of the twentieth century' \cite{dongarraGuestEditorsIntroduction2000, limTensorsComputations2021} or the recent conceptual role of tensors in the ML discovery of faster matrix multiplication algorithms \cite{fawziDiscoveringFasterMatrix2022}.  
The ubiquitous utility of tensors and tensor-based methods in computational applications has inspired systems architects explore hardware design solutions for tensor-based computations similar to the matrix-based motivation behind GPU and TPU such as a recently proposed hardware accelerator for sparse and dense tensor factorization \cite{srivastavaTensaurusVersatileAccelerator2020}.

\subsection{Motivation and Research Questions} \label{sec:Intro:RQs}
The purpose of this paper is to review current state-of-the-art methods for DC as well as to motivate and illustrate how the properties of tensors can be leveraged to improve DC. 
While the success of methods in TDA are built upon an ability to capture higher-order characteristics, to our knowledge there are limited works using TDA specifically for the purpose of DC.
This survey addresses the following research questions (RQs).
\begin{itemize}
    \item RQ1: What properties of tensors and TDA methods are well-suited for the DC problem space?
    \item RQ2: To what extent has TDA been explored in the problem space of DC in existing studies?
    \item RQ3: What are the canonical DC methods, e.g. structural, statistical, and model-based methods, and how do TDA methods align with these canonical DC methods?
    \item RQ4: What contemporary problem spaces could benefit from improved tensor based DC?
\end{itemize}

\subsection{Paper Organization}
Section \ref{sec:DC} discusses the general practice of DC and provides illustrative examples.  Meta-features are presented as a high level example of applied DC along with a framework for comparative analysis of datesets that incorporates multiple types of characteristics (statistical, structural, and performance-based) is presented as alternate view of the nuances of DC in applications.  
Section \ref{sec:Tensors} provides background information and notation details for tensors and tensor data analysis, a review of existing tensor surveys, and a review of existing works in TDA with potential application for the characterization of datasets in terms of statistical, structural, and model-based properties. 
Section\ref{sec:Illustrative Examples} presents two examples that illustrate properties of tensors and tensor methods that are useful for the sake of DC, malleable problem formulations and implicit solutions: a generalized form of the Canonical Polyadic tensor decompostion and a novel approach for the comparison of tensors.
Lastly, RQs are summarized, analyzed, and discussed in Section \ref{sec:Conclusion} as a conclusion.

\section{Dataset Characterization: Background and Applications}
\label{sec:DC}
DC consists of a group of measures calculated directly from a dataset with the intent to illuminate intrinsic properties so as to facilitate analysis and computation.
The history of DC spans back to the advent of databases and statistical software where efforts were primarily focused on applying statistical methods and measures like central tendency and dispersion to better understand datasets.   
DC was elaborated upon in the late 1970s and early 1980s by John Tukey and others through the development of exploratory data analysis (EDA) \cite{tukeyExploratoryDataAnalysis1977} which incorporates summary details and visualizations to augment traditional statistical hypothesis testing.  
The EDA community laid the foundations for modern data mining and developed techniques like dimensionality reduction and ordination.  
Efforts in DC were further advanced by the rise of ML in the 1990s with data-centric algorithms for tasks like classification, prediction, and clustering that can provide characterizations based on more complex feature spaces and continue in the current era of big data and deep learning.
Modern DC considerations include: descriptive statistics, data quality assessments, visualization, feature analysis, dimensionality, structure, pattern recognition, model-based characterization,  and the comparative analysis of datasets.

\subsection{Meta-Features in Meta-Learning} \label{sec:Meta-Features}
DC plays a critical role in meta-learning, an area of ML that aims to simulate the human ability to leverage prior knowledge when faced with a new task.  
This process relies heavily on meta-features (i.e. dataset characteristics) calculated using a multitude of concepts that can be distinctly grouped. 
The algorithm selection problem, formalized by Rice \cite{riceAlgorithmSelectionProblem1976}, is a canonical problem in meta-learning that uses meta-features to determine which candidate algorithm will perform best on a given dataset and has given rise to a wide variety of types of dataset characteristics to develop improved solutions.
Meta-features tend to lack a uniform definition, but can be categorized into specific types.
\subsubsection{Types of Meta-features}
Simple measures are derived directly from a dataset, e.g.  number of attributes, classes, or missing entries.
Statistics based measures reflect statistical properties of a dataset, e.g. mean, standard deviation ratio, skewness, or kurtosis \cite{engelsUsingDataMetric1998, sohnMetaAnalysisClassification1999}.
Information-theoretic measures are typically calculated based on entropy measures of a dataset, e.g. entropy of classes or mean mutual information of classes and attributes based on nominal attributes \cite{segreraInformationTheoreticMeasuresMetalearning2008}.
Landmarking measures are based on how a set of simple classification algorithms (referred to as landmarker) perform on a dataset, e.g. accuracy \cite{bensusanCasaBatloPasseig2000, bernhardMetalearningLandmarkingVarious2000}.
Model-based measures build a model, typically a decision tree, with a dataset and use properties of the model to characterize the data, e.g. depth of tree, number of nodes, or number of leaves \cite{bensusanHigherorderApproachMetalearning2000, pengImprovedDatasetCharacterisation2002}.
Problem complexity measures are used to characterize classification datasets by analyzing the sources of difficulty in solving a classification task and are designed to reflect the way different classes are separated or interleaved, e.g. Fischer's discriminant ratio or nonlinearity of linear classifier \cite{basuDataComplexityPattern2006, smithInstanceLevelAnalysis2014}.
More recent work employ feature vectors based on frequencies of itemsets with respect to a function (parity or conjunction) and clustering \cite{songAutomaticRecommendationClassification2012, wangImprovedDataCharacterization2015}.

\subsection{Comparative Analysis} \label{sec:Dataset Characterization:ComparativeAnalysis}
Comparative analysis of datasets relies on a systematic comparison of myriad dataset characteristics. 
Robnik-$\breve{\text{S}}$ikonja  \cite{robnik-sikonjaDatasetComparisonWorkflows2018} proposes a general framework for the comparative analysis of datasets that affords the inclusion of multiple dataset characteristics.
As presented, the framework considers statistical, structural, and model-based measures, but is general in the sense that any multitude of measures or class of measures maybe incorporated.
This work highlights the complexity and nuances of dataset comparison and how a variety of views of a dataset can be used to improve discriminatory results and insights.

\subsubsection{Statistical Comparison} \label{stats_comp}
Comparison of two low dimensional datasets commonly relies on statistics for each attribute based on central moments: mean, median, standard deviation, skewness, and kurtosis.  
Statistical moments can offer important insights into the data, but become insufficient as the number of attributes present increases.  
Additionally, summary statistics fall short of providing a global view of the data as they can fail to reveal possible interactions between attributes.   

\begin{figure}[h]
  \centering
  \includegraphics[width=.9\columnwidth]{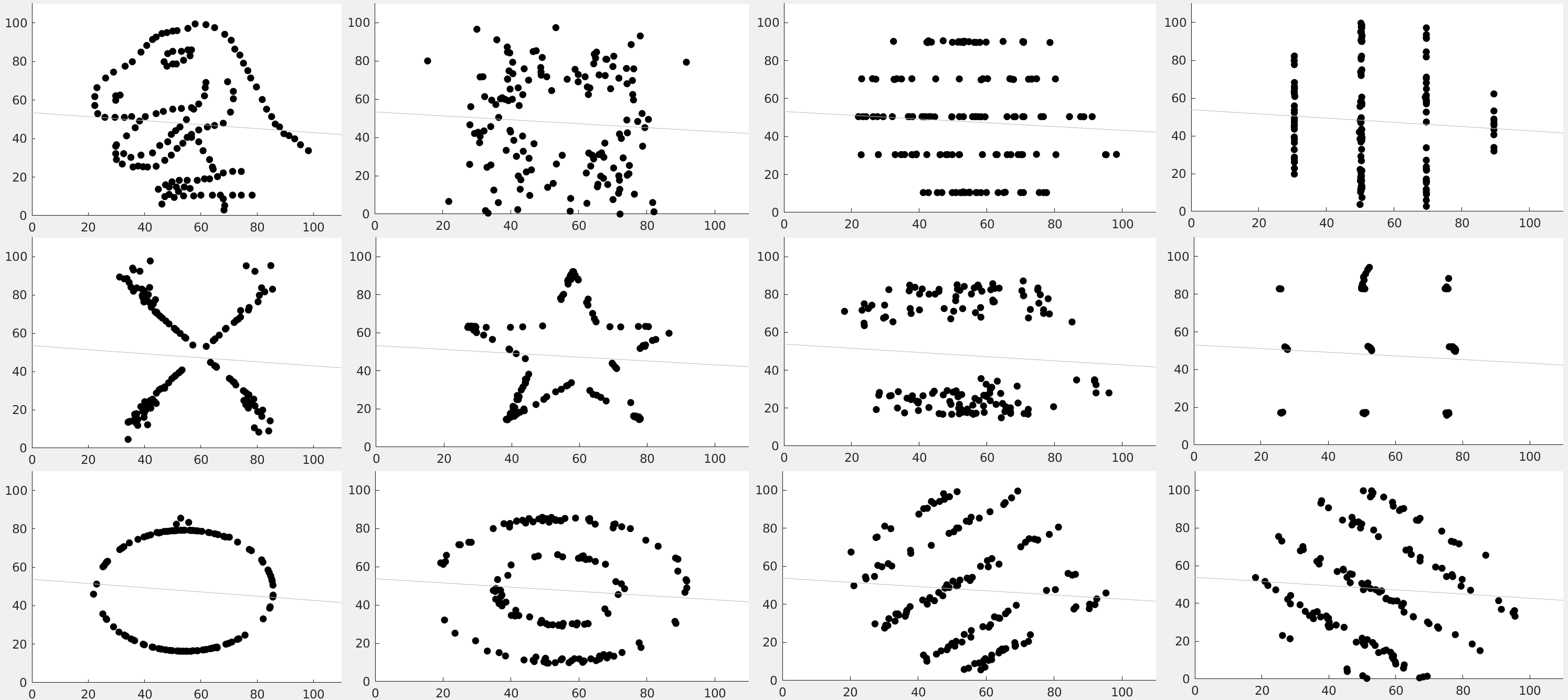}
  \caption{Sampling of the Datasaurus Dozen.  Each dataset share the same summary statistics to two decimal places.}
  \label{datasaurus}
\end{figure}

The Datasaurus Dozen \cite{matejkaSameStatsDifferent2017} is an salient example of how comparison of summary statistics alone can be misleading that modernizes the canonical Anscombe's quartet \cite{anscombeGraphsStatisticalAnalysis1973}.  
The authors of \cite{matejkaSameStatsDifferent2017} used simulated annealing to systematically adjust the dataset in the upper left subgraph of Figure \ref{datasaurus} to generate additional datasets with shared statistics that are clearly different upon visualization.

There are additional drawbacks to using summary statistics. 
For example, skewness and kurtosis are difficult to compare across multiple attributes as they require a symmetric distribution for interpretation, are sensitive to outliers, and have values spanning $-\infty$ to $\infty$.  
\cite{robnik-sikonjaDatasetComparisonWorkflows2018} suggests the inclusion of the more robust variants medcouple (MC) and left/right medcouple (L/RMC), introduced in Brys et al \cite{brysRobustMeasuresTail2006a} which are independent of the mean and standard deviation which makes them resilient to outliers. 

\begin{figure}[h]
  \centering
  \includegraphics[width=.9\columnwidth]{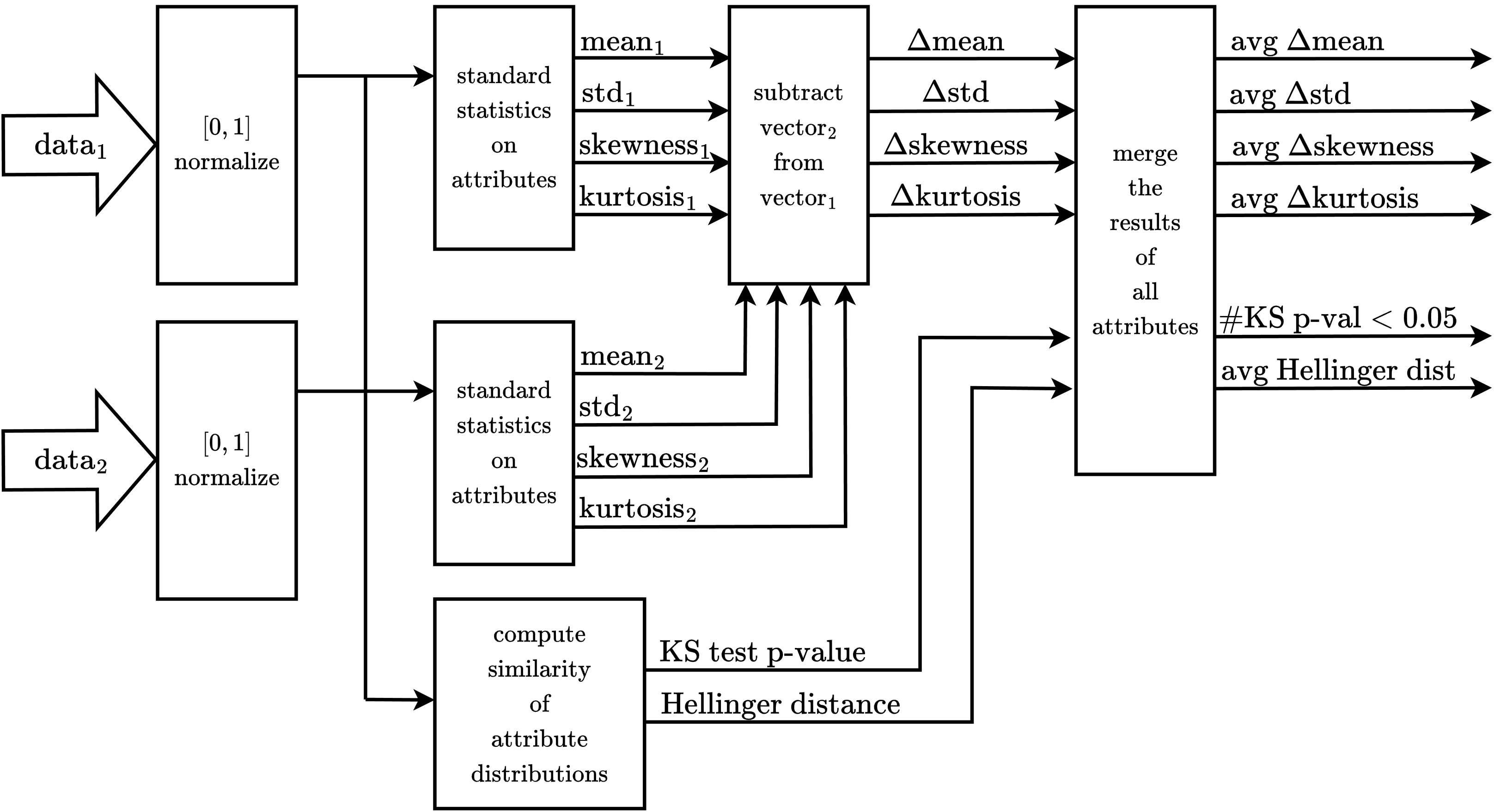}
  \caption{Workflow for comparing standard statistics between two datasets \cite{robnik-sikonjaDatasetComparisonWorkflows2018}}
  \label{DSCW_fig:2}
\end{figure}

Figure \ref{DSCW_fig:2} illustrates the statistical comparison workflow proposed by \cite{robnik-sikonjaDatasetComparisonWorkflows2018}.  
Numerical attribute values are normalized to facilitate comparison across attributes.  
The average difference for each statistic across all attributes is reported a measure of similarity.
Hellinger distance is used to compare distributions of discrete attributes.  
Alternate measures for comparing the discrete attributes include Kullback-Leibler distance, Bhattacharyya distance, and Jensen-Shannon distance.  
A two sample Kolomgorov-Smirnov (KS) test is used for numerical attributes to test the difference between two probability distributions.
Alternate distance measures for comparing numerical attributes include the Cucconi test \cite{cucconiNuovoTestNon1968} and the Lepage test \cite{lepageCombinationWilcoxonAnsariBradley1971}.
Smaller values for both the Hellinger distance and KS test indicate greater similarity.

Choice of measures for both numerical and discrete attributes emphasize the adaptability of the statistical comparison workflow to incorporate domain knowledge associated with a dataset. 

\subsubsection{Clustering Comparison} \label{clustering_sim}
Clustering algorithms partition dataset instances into similar groups to provide information about the structure of the data.  
For instance, clustering analysis would immediately differentiate between datasets in the Datasaurus Dozen example \cite{matejkaSameStatsDifferent2017}.
There are two main categories of clustering validation techniques: internal and external.  
Internal validation relies exclusively on partition data to assess a partitioning for a given purpose.  
It considers the balance between cluster compactness and inter-cluster separation, both of which may be defined in a multitude of ways.  
External validation evaluates the degree of matching between competing clusterings.  
Theoretical results indicates that no single clustering similarity index outperforms all others \cite{jaskowiakStrategiesBuildingEffective2016} as a consequence of many clustering measures being strongly correlated \cite{vinhInformationTheoreticMeasures2010}. 

While existing internal and external cluster validation methods target clusterings of a single dataset, \cite{robnik-sikonjaDatasetComparisonWorkflows2018} presents a novel approach to clustering that allows for comparison across multiple datasets.

\paragraph{Similarity of two clusterings on the same dataset}

Given the dataset $D = \{ (\textbf{x}_i)_{i=1}^N ) \}$ with $N$ data points.  
Assume two clusterings of $D$, $U = \{ U_1,\dots,U_u  \}$ and $V = \{ V_1, \dots, V_v \}$, where $U_1 \cap \dots \cap U_u = \emptyset $, $U_1 \cup \dots \cup U_u = D$, $V_1 \cap \dots \cap V_v = \emptyset$, and $V_1 \cup \dots \cup V_v = D$.  
The information overlap of the clusters $U$ and $V$ can be expressed with a $ u \times v$ contingency table, see Table \ref{DSCW_tbl:1}.

\begin{table}[htbp]
  \caption{The contingency table of clusterings overlap, $n_{i,j} = | U_i \cap V_j |$}
  \label{DSCW_tbl:1}
  \begin{center}
  \begin{tabular}{|c|cccc|c|}
    \hline
    $U/V$ & $V_1$ & $V_2$ & \dots & $V_v$ & Sum\\
    \hline
    $U_1$ & $n_{1,1}$ & $n_{1,2}$ & \dots & $n_{1,v}$ & $u_1$ \\
    $U_2$ & $n_{2,1}$ & $n_{2,2}$ & \dots & $n_{2,v}$ & $u_2$ \\
    $\vdots$ & $\vdots$ & $\vdots$ & $\ddots$ & $\vdots$ & $\vdots$ \\
    $U_u$ & $n_{u,1}$ & $n_{u,2}$ & \dots & $n_{u,v}$ & $u_u$ \\
    \hline
    Sum & $v_1$ & $v_2$ & \dots & $v_v$ & $N$ \\
    \hline
  \end{tabular}
  \end{center}
\end{table}

Vinh et al. \cite{vinhInformationTheoreticMeasures2010} presents several measures for comparing clustering based on counting pairs of points on which two clusterings agree or disagree.  
Each of the $\genfrac{(}{)}{0pt}{1}{N}{2}$ distinct pairs of data points in $D$ falls into one of four categories:
$\mathbf{N_{11}}$ - number of pairs in the same cluster in both $U$ and $V$,
$\mathbf{N_{00}}$ - number of pairs in the different clusters in both $U$ and $V$,
$\mathbf{N_{01}}$ - number of pairs in the same cluster in $U$ and different clusters in $V$, or
$\mathbf{N_{10}}$ - number of pairs in different cluster in $U$ and in the same cluster in $V$.

The values $\mathbf{N_{11}}$ and $\mathbf{N_{00}}$ indicated the degree to which clusterings $U$ and $V$ overlap, while the values $\mathbf{N_{01}}$ and $\mathbf{N_{10}}$ indicate the degree by which the clusterings differ.  
These values form the basis for external clustering comparison metrics like the Rand Index (RI) \cite{randObjectiveCriteriaEvaluation1971}.  
RI, defined as,
\begin{equation}\label{DSCW_eq:RI}
    RI(U,V) = \frac{\mathbf{N_{00}} + \mathbf{N_{11}}}{{n \choose 2}}.
\end{equation}
lies between 0 and 1, where a value of 1 indicates two clusterings are identical and a value of 0 indicates no pair of points appears in either the same cluster or different clusters in $U$ and $V$.  
The RI counts pairs of elements in both $\mathbf{N_{11}}$ and $\mathbf{N_{00}}$ and can be adversely affected as the number of clusters increase.
The Adjusted Rand Index (ARI) metric was proposed by Hubert and Arabie \cite{hubertComparingPartitions1985} to correct for chance and provide a similarity metric that takes a value close to zero for two random clusterings (not afford by the RI).  
The ARI accomplishes this by employing a generalized hypergeometric distribution to model randomness and computes the expected number of entries in the contingency table.
\begin{equation}\label{DSCW_eq:ARI}
    ARI = \frac{RI - E[RI]}{\max RI - E[RI]} \\
\end{equation}
Note, the ARI has an expected value of 0 for a random distribution of clusters, a value of 1 when clusters perfectly match, and negative values when the amount of overlap is less than expected.
Additional clustering metrics include Fowlkes-Mallows (FM) \cite{fowlkesMethodComparingTwo1983}, Jaccard (J) index \cite{jaccardDistributionCompareeFlore1901}, and variation of information (VI) \cite{meilaComparingClusteringsInformation2007}. 

\paragraph{Similarity of two clusterings on different datasets}

\begin{figure}[h]
  \centering
  \includegraphics[width=.9\columnwidth]{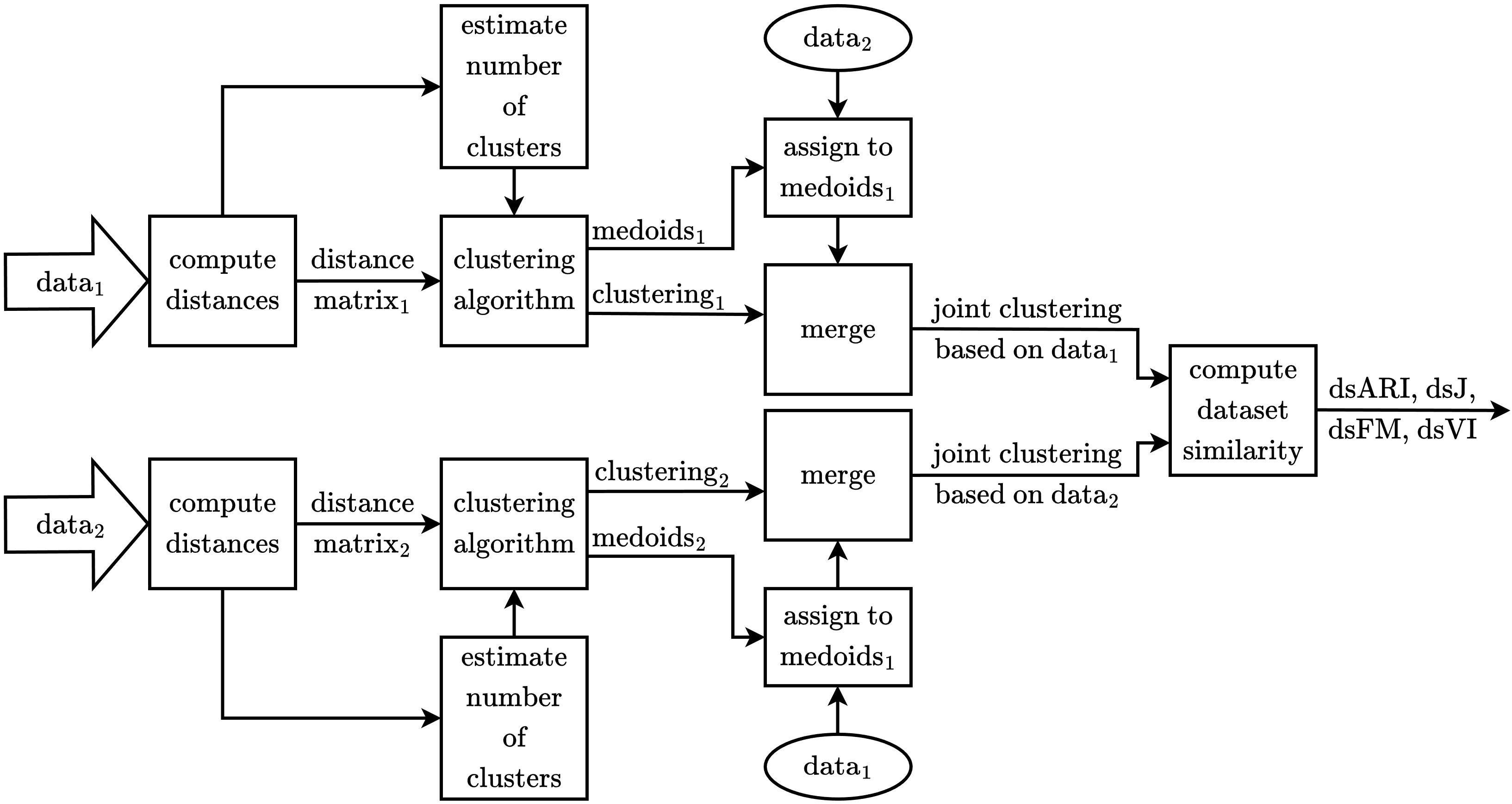}
  \caption{Workflow for comparing clustering similarity between two datasets \cite{robnik-sikonjaDatasetComparisonWorkflows2018}}
  \label{DSCW_fig:3}
\end{figure}

In order to extend clustering comparison to \emph{two} datasets, \cite{robnik-sikonjaDatasetComparisonWorkflows2018} first clusters each dataset separately and extracts each cluster medoid (i.e. the data instance within a cluster with maximal average similarity to all other instances in the cluster).
Joint clusterings are defined on both datasets where each instance in one datasets is compared to the other dataset's medoids and assigned to the cluster of the medoid the instance is nearest too in order to access standard external clustering similarity measures.  
The result is a clustering of the union of the datasets based on the structure of just one of the datasets.  
See Figure \ref{DSCW_fig:3} for an flowchart illustrating the process described above.

For convenience \cite{robnik-sikonjaDatasetComparisonWorkflows2018} makes use of the partitioning around medoids (PAM) clustering algorithm 
which provides partitions and mediods as its output \cite{kaufmanFindingGroupsData2005}.  
Distance to the medoids is the criteria used for assigning new instances to existing clusters and is computed by Gower's method \cite{gowerGeneralCoefficientSimilarity1971}, which normalizes numerical attributes to $[0,1]$ and employs a 0-1 scoring of dissimilarity between nominal attributes where the distance is given as the sum of dissimilarities over all attributes.
The similarity index (e.g. ARI or FM) is computed once the joint clusterings are formed.  
Other options for clustering, assignment criteria, estimating optimal number of clusters, and standard measures may be employed as dictated by domain.

A thoughtful construction of clustering comparison across datasets like the workflow proposed by \cite{robnik-sikonjaDatasetComparisonWorkflows2018} provides insights into the structural characteristics that can be used in tandem with the statistical and performative characteristics of a dataset for more meaningful comparisons.

\subsubsection{Model-Based Comparison} \label{perfom_comp}
Performance of classification models round out the dataset comparison workflows proposed by Robnik-$\breve{\text{S}}$ikonja  \cite{robnik-sikonjaDatasetComparisonWorkflows2018}. 
Model performance on tasks can be used as a measure of dataset characteristics as exemplified by the aforementioned landmarking and model-based meta-features used in meta-learning.
Classification is a primary task in the realms of data mining and ML and is suitable for exploring dataset interchangeability and dataset fusion, both of which require dataset comparison.  
\begin{figure}[h]
  \centering
  \includegraphics[width=.9\columnwidth]{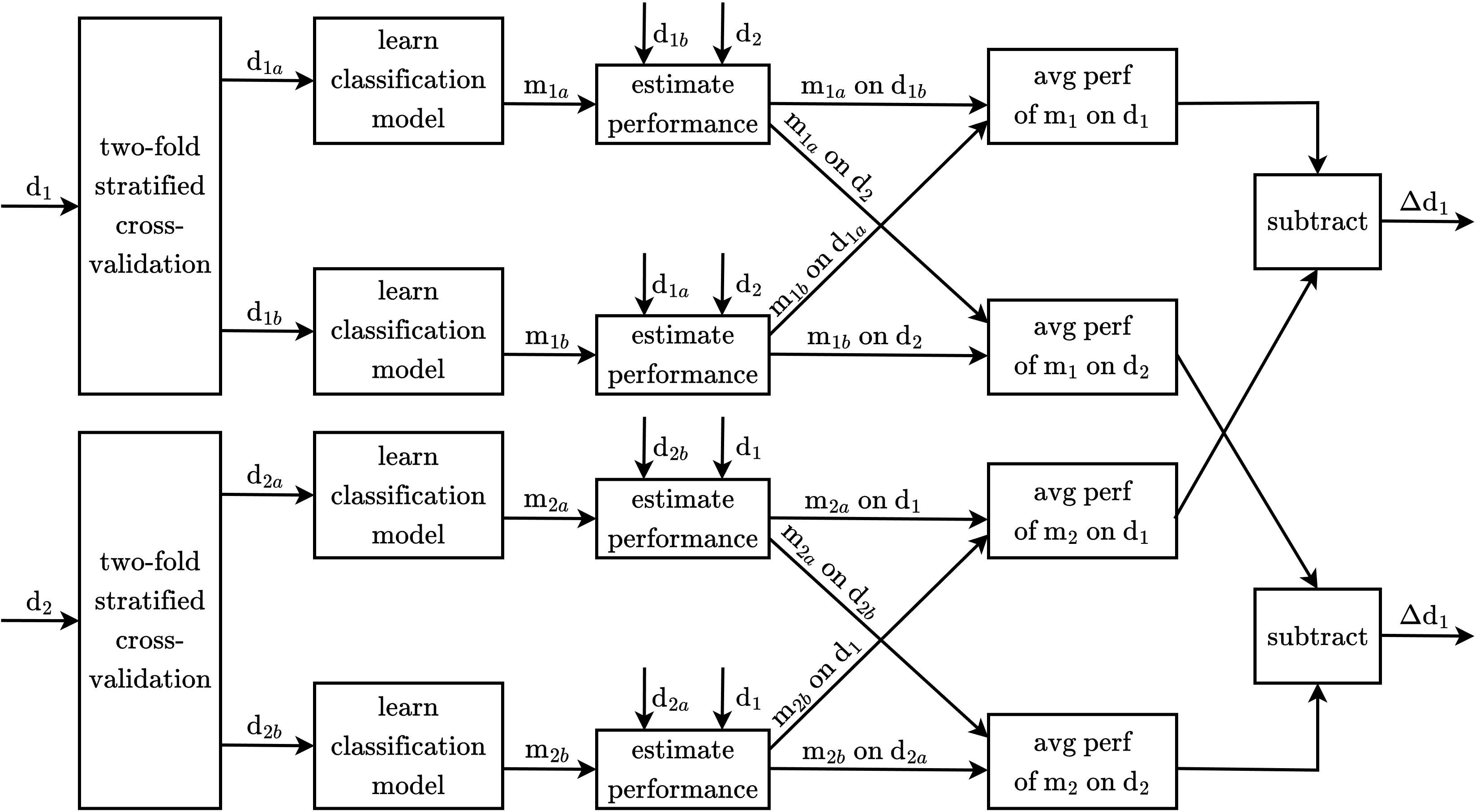}
  \caption{Workflow for comparing classification performance between two datasets \cite{robnik-sikonjaDatasetComparisonWorkflows2018}}
  \label{DSCW_fig:4}
\end{figure}
The approach, see Figure \ref{DSCW_fig:4}, involves training a classification model for each dataset exclusively. 
Each of the models are then tested with unseen data from both datasets to achieve a performance comparison.
Through this lens, a model trained on one dataset that performs comparably on both datasets indicates the dataset not used in training lies within the distribution of the training dataset and is a suitable substitute.  
Similarly, if the trained model performs better on the dataset not used for training, that is an indication that the dataset is a simplified version of the training dataset.  

\cite{robnik-sikonjaDatasetComparisonWorkflows2018} use a stratified split for two datasets, $d_1$ and $d_2$, into four datasets, $\{d_{1a}$, $d_{1b}$, $d_{2a},$ $d_{2b}\}$, to train classification models, $\{m_{1a}, m_{1b}, m_{2a},$ $m_{2b}\}$, (i.e. two models trained per dataset, each trained on different data).  
Performance across the models is tested using all unseen data, i.e. $m_{1a}$ is tested using $d_{1b}$ and all of $d_2$.  
The resulting performance scores allow model performance estimates for both models, $m_1$ and $m_2$, on each dataset, $d_1$ and $d_2$.
The difference of model performances, $\Delta d_1$ and $\Delta d_2$, is considered as the best indicator of similarity between the datasets in terms of the classification task, i.e. a $\Delta d_1$ close to zero indicates that $d_2$ would be a suitable substitute for $d_1$ and the model learned using $d_2$ ought to achieve comparable performance on $d_1$.

Performative comparisons as described can be an informative and effective technique for DC, especially when the comparison involves how the data will be used, and have the capacity to capture the presence of higher-order correlations depending on the model employed.
A downside to performance based comparison is that it is more computationally and resource intensive than  statistical or structural comparison.

\section{Tensors and Tensor Data Analysis: Background and Existing Survey Review}
\label{sec:Tensors}  
Tensors, first formalized in the development of differential geometry by Gauss, Riemann, and Christoffel and made broadly accessible by Ricci and Levi-Civita'  \cite{ricciMethodesCalculDifferentiel1900}, have enjoyed resurgent interest in the 21st century in the area of computing and large-scale data analysis.
This is owed in part to the tensor's suitability for multi-dimensional data and the multilinear algebra attached that affords more rigorous theoretical guarantees than matrices \cite{koldaTensorDecompositionsApplications2009,domanovUniquenessComputationDecomposition2020, kruskalThreewayArraysRank1977}.
The chronological evolution of the tensor is embodied by the following essential definitions,
\begin{enumerate}
    \item as a multi-indexed object that \emph{satisfies certain transformation rules},
    \item as a multilinear map,
    \item as an element of a tensor product of vector spaces,
\end{enumerate}
where the latter two, more modern, definitions directly encode the transformation rules of the first \cite{limTensorsComputations2021}.  

The mathematical definition(s) of tensors coupled with their ability to represent multi-dimensional data have led to their broad proliferation in myriad problem spaces involving large-scale data analysis.
See Section \ref{sec:Tensors:ExistingSurveys} for review of existing surveys of tensor based applications, including: ML, signal processing, numerical linear algebra, and graph analysis.

\subsection{Basic Notation, Terminology, and Operations for Tensors}

The following conventions are adhered to.
Scalar values are given as a lower case letter, $x$, vectors as a lower case boldface letter, $\textbf{x}$, matrices as an upper case boldfaced letter, $\textbf{X}$, and tensors as a boldface uppercase Euler font letters, $\boldsymbol{\mathcal{X}}$.  
The number of dimensions, determines the \emph{mode} or \emph{order} of a tensor.
Thus, a scalar is an mode-0 tensor, a vector is an mode-1 tensor, a matrix is an mode-2 tensor, a data cube is an mode-3 tensor, and generalizes to some mode-$d$ tensor $\boldsymbol{\mathcal{X}} \in \mathbb{R}^{I_1 \times I_2 \times \cdots \times I_d}$. 

Elements of a tensor are indicated using a multi-index notation, i.e. $x_i \equiv x_{i_1,\dots, i_d}$ denotes an entry $i = (i_1, \dots, i_d) \in \mathcal{I} \equiv \{ 1, \dots, I_1\} \bigotimes \cdots \bigotimes \{1,\dots,I_d\}$ of an mode-$d$ tensor, $\boldsymbol{\mathcal{X}} \in \mathbb{R}^{I_1 \times I_2 \times \cdots \times I_d}$ where $\mathcal{I}$ is the set of all possible combinations of indices.  

Fibers of a tensor are defined by fixing all but one index of a tensor and generalize the concept of columns and rows in matrices.
Vectorization rearranges the elements of the tensor $\boldsymbol{\mathcal{X}}$ by stacking the $k$-mode fibers into a single vector, $\textbf{x} = \vect(\boldsymbol{\mathcal{X}})$.  
Slices of a tensor are defined by fixing all but two indices of a tensor, resulting in a sub-matrix of a tensor.
For example, the $k$-th frontal slice of $\boldsymbol{\mathcal{X}} \in \mathbb{R}^{I_1 \times I_2 \times I_3}$ is $\textbf{X}^{(k)} = \boldsymbol{\mathcal{X}}(:,:,k)$

\begin{figure}[h]
  \centering
  \includegraphics[height=1.5in]{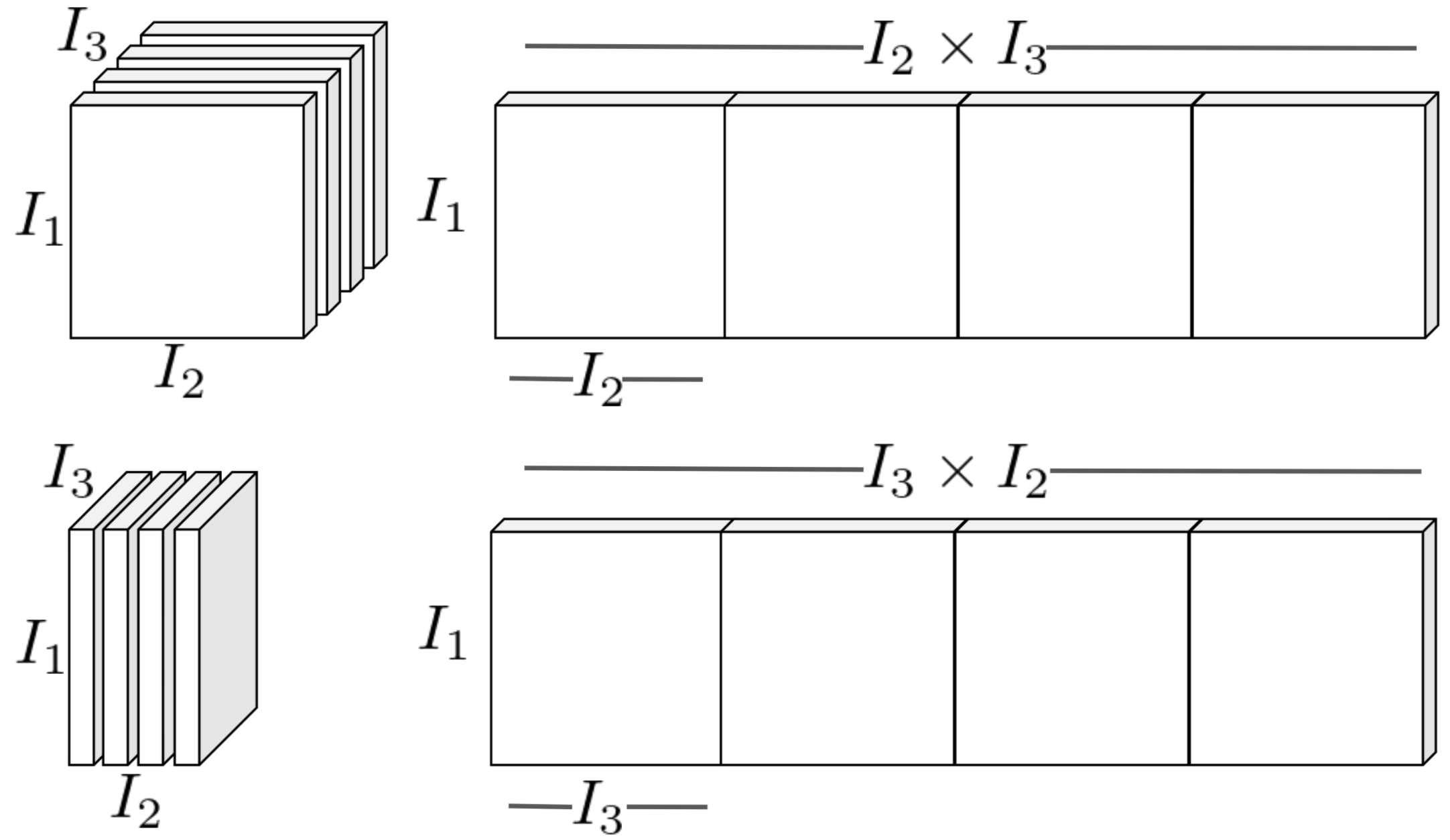}
  \caption{Forward (top) and Backward (bottom) Mode-1 matricization of a tensor.}
  \label{Tensors_fig:Matricization}
\end{figure}

Matricization rearranges the elements of a tensor $\boldsymbol{\mathcal{X}}$ by stacking slices, relative to specific  modes, into a single matrix, and can take many forms.
Figure \ref{Tensors_fig:Matricization} illustrates the forward and backward mode-1 unfoldings of an mode-3 tensors, i.e. 
$\textbf{X}_{(k)} \in \mathbb{R}^{I_k \times \frac{I_1\cdots I_d}{I_k}}$.    
The authors of \cite{domanovComputationComparisonTensor2021} employ a mode-$n$ matrix representation $\textbf{X}_{(n^c;n)} \in \mathbb{R}^{\frac{I_1\cdots I_d}{I_n} \times I_n}$ where the columns of the matrix are the vectorized mode-$n$ slices of $\boldsymbol{\mathcal{X}}$.

\subsubsection{Multilinear Algebra and Tensor Data Analysis Basics}
Tensors are central to multilinear algebra, a branch of mathematics that extends linear algebra to the study of multilinear maps and the structures that arise from transformations of spaces, as they allow for the representation and preservation of higher-dimensional relationships between algebraic objects that are more complex than those captured by linear algebra.
Similar to matrices and vectors, tensors are equipped with algebraic operations like tensor addition, scalar multiplication, inner and outer products.
Many operations like tensor multiplication require a more complex set of notation and symbols and can take a variety of forms, see Bader and Kolda \cite{baderAlgorithm862MATLAB2006,baderEfficientMATLABComputations2008} for thorough treatment.

The $n$-mode product is a type of tensor multiplication where a tensor is multiplied by a matrix or vector in the $n$-th mode.
For a tensor, $\boldsymbol{\mathcal{X}}\in\mathbb{R}^{I_1 \times \cdots \times I_d}$, and a matrix,$\textbf{U}\in\mathbb{R}^{J\times I_n}$, the $n$-mode product is denoted as:
\begin{equation}\label{eq:n-mode product}
    \boldsymbol{\mathcal{Y}} = (\boldsymbol{\mathcal{X}}\times_n \textbf{U})_{i_1 \dots i_{n-1}ji_{n+1}\dots i_d} = \sum_{i_n=1}^{I_n}x_{i_1i_2\dots i_d}u_{ji_n},
\end{equation}
where $\boldsymbol{\mathcal{Y}} \in \mathbb{R}^{I_1 \times \cdots I_{n-1} \times J \times I_{n+1} \times \cdots \times I_{d}}$.
This operation is sometimes referred to as the mode-$n$ inner product and denoted with the symbol $\bullet_n$.

For a series of products, order of multiplication is arbitrary when along the distinct modes,
\begin{equation}
    \boldsymbol{\mathcal{X}} \times_m \textbf{A} \times_n \textbf{B} = \boldsymbol{\mathcal{X}} \times_n \textbf{B} \times_m \textbf{A}
\end{equation}
but must be considered when along the same mode
\begin{equation}
    \boldsymbol{\mathcal{X}} \times_n \textbf{A} \times_n \textbf{B} = \boldsymbol{\mathcal{X}} \times_n (\textbf{BA}).
\end{equation}

The Einstein product is a generalized product of tensors that defines multiplication across multiple modes and reduces to standard matrix multiplication and the $n$-mode product as special cases. 
For a tensor $\boldsymbol{\mathcal{X}}\in\mathbb{R}^{I_1 \times \cdots \times I_M \times J_1 \times \cdots \times J_N}$ and a tensor $\boldsymbol{\mathcal{Y}}\in\mathbb{R}^{J_1 \times \cdots \times J_N \times K_1 \times \cdots \times K_L}$, the Einstein product is denoted as:
\begin{equation}
    \begin{split}
    \boldsymbol{\mathcal{Z}} &= (\boldsymbol{\mathcal{X}}\star_N \boldsymbol{\mathcal{Y}})_{i_1,\dots,i_{M},k_1,\dots.k_{L}\dots i_d} \\ 
    &= \sum_{j_1\dots j_N}x_{i_1,\dots i_M,j_1,\dots,j_N}y_{j_1,\dots,j_N,k_1,\dots,k_L}.
    \end{split}
\end{equation}
where $\boldsymbol{\mathcal{Z}} \in \mathbb{R}^{I_1 \times \cdots \times I_{M} \times K_1 \times \cdots \times K_{L}}$.

Many TDA methods rely on finding a solution to systems of multilinear equations (akin to a system of linear equations) which can be formulated as a tensor equation using the Einstein product \cite{changMultiRelationalDataCharacterization2021}.
\begin{equation} \label{eq:multilinearsystemofeqs}
    \textbf{Ax} = \textbf{b} \quad \Rightarrow \quad
    \boldsymbol{\mathcal{A}} \star_N \boldsymbol{\mathcal{X}} = \boldsymbol{\mathcal{B}}.
\end{equation}

\subsubsection{Tensor Measures}
Tensors are equipped with various measures of 'size' at the theoretical level, which is useful for the sake of DC.
The most notable are tensor norms and tensor ranks, which play an indispensable auxiliary role in many TDA methods.

The norm of a tensor provides a way to quantify the magnitude of its components.
The generally accepted convention for the \emph{norm} of a tensor is to use the Forbenius norm (i.e. the higher-order equivalent of the Euclidean norm),
\begin{equation} \label{eq:Forbenius Norm}
    \| \boldsymbol{\mathcal{X}} \|_{\text{F}} = 
    \sqrt{\langle \boldsymbol{\mathcal{X}},\boldsymbol{\mathcal{X}} \rangle} =
    \sqrt{\sum_{i_1 = 1}^{I_1} \sum_{i_2 = 1}^{I_2} \cdots \sum_{i_N = 1}^{I_N} x_{i_1 i_2 \dots i_N}^2},
\end{equation}
and is commonly designated as $\| \boldsymbol{\mathcal{X}} \|$. 
Other tensor norms include the $L_1$ norm given as the sum of the absolute values of all the elements in the tensor which provides a measure of sparsity or the $L_{\infty}$ norm given as the maximum absolute value of the elements in the tensor. 
The spectral norm \cite{chenTensorSpectralPnorm2020, friedlandSpectralNormSymmetric2020} and nuclear norm \cite{signorettoNuclearNormsTensors2011} are more complex and typically defined in the context of specific applications.

The rank of a tensor is a concept used to measure the complexity of a tensor and is analogous to the rank of a matrix, although it is not uniquely defined.
In the matrix case, $rank(\textbf{A})$, is the maximum number of linearly independent columns (or rows) in a matrix and measures the dimension of the vector spaced spanned by the columns (or rows) of a matrix.

The mode-$k$ rank of a tensor is the generalization of the column or row rank of a matrix and is defined as the dimension of the subspace spanned by the mode-$k$ unfolding of a tensor $\boldsymbol{\mathcal{X}}$, $rank(\textbf{X}_{(k)})$. 
The multilinear rank  of a tensor is the $d$-tuple consisting of mode-$k$ rank for all $k$, i.e. a third-order tensor has a multilinear rank of $(rank(\textbf{X}_{(1)}),rank(\textbf{X}_{(2)}),rank(\textbf{X}_{(3)}))$. 
A rank-1 tensor has its mode-$k$ rank equal to one for all $k$ and is formed from the outer product of $k$ non-zero vectors.  
The minimal number of rank-1 tensor required in linear combination to yield a tensor $\boldsymbol{\mathcal{X}}$ is referred to as the tensor rank or CP rank.  
See \cite{koldaTensorDecompositionsApplications2009,comonGenericTypicalRanks2009} for thorough treatments of tensor rank and the subtleties associated with the concept.
It is known that determining the rank of a tensor is NP-complete\cite{hastadTensorRankNPcomplete1990}.

\subsubsection{Tensor Expressions in Matrix Form}
Tensor expressions can be reformulated implicitly as matrix problems via tensor manipulations and a variety of matrix products.
For instance, The $n$-mode product can be easily expressed in terms of an unfolded tensor and matrix multiplication via the following relationship:
\begin{equation} \label{eq:matricized mode-n product}
    \boldsymbol{\mathcal{Y}} = \boldsymbol{\mathcal{X}}\times_n\textbf{U} \Leftrightarrow \textbf{Y}_{(n)} = \textbf{UX}_{(n)}.
\end{equation}
The most commonly encountered products  are the \emph{Kronecker product}, the \emph{Khatri-Rao product} (KRP), and the \emph{Hadamard product}.
The Kronecker product of two matrices, $\textbf{A}\in\mathbb{R}^{k\times l}$ and $\textbf{B}\in\mathbb{R}^{m\times n}$, scales the matrix $\textbf{B}$ by each element of $\textbf{A}$ where $\textbf{A} \otimes \textbf{B} \in \mathbb{R}^{km\times ln}$.
The KRP is equivalent to a column-wise Kronecker product where the elements of $\textbf{A}$ scale the columns of $\textbf{B}$ such that for matrices $\textbf{A}\in\mathbb{R}^{m\times p}$ and $\textbf{B}\in\mathbb{R}^{n\times p}$, $\textbf{A} \odot \textbf{B}\in \mathbb{R}^{mn\times p}$.
The Hadamard product is an element-wise multiplication of two same-sized matrices $\textbf{A,B} \in \mathbb{R}^{m\times n}$ where $\textbf{A}* \textbf{B} \in \mathbb{R}^{m\times n}$

For example, Kolda \cite{koldaMultilinearOperatorsHigherorder2006} provides a proof of the following usage of the Kronecker product for expressing a series of mode-$n$ products as a matrix calculation where for a tensor $\boldsymbol{\mathcal{X}}\in\mathbb{R}^{I_1\times I_2\times\dots\times I_N}$ and matrices $\textbf{A}^{(n)}\in \mathbb{R}^{J_n\times I_n}$, then for any $n \in \{1,\dots,N\}$

\begin{equation}
    \scalebox{0.8}{$
    \begin{aligned}
    \boldsymbol{\mathcal{Y}} &= \boldsymbol{\mathcal{X}}\times_1\textbf{A}^{(1)}\times_2\textbf{A}^{(2)}\dots\times_N\textbf{A}^{(N)} \Leftrightarrow \\
    \textbf{Y}_{(n)} &= \textbf{A}^{(n)}\textbf{X}_{(n)}\left(\begin{matrix}\textbf{A}^{(N)}\otimes\dots\otimes\textbf{A}^{(n+1)}\otimes\textbf{A}^{(n-1)}\otimes\dots\otimes\textbf{A}^{(1)}\end{matrix}\right)^{\mathsf{T}}.
    \end{aligned}
    $}
\end{equation}

\subsection{Tensor Data Analysis: Applied Multilinear Algebra}
The multilinear algebra attached to tensors allows for a set of methods in TDA that includes: tensor decomposition, tensor networks, tensor completion, tensor clustering, tensor regression and classification, multilinear subspace learning, tensor-based anomoly detection, and dynamic tensor analysis.

\subsubsection{Tensor Decomposition: TDA Workhorse}
Tensor decompositions factorize a tensor into products and/or sums of simpler tensors, matrices, and vectors. 
The form of a tensor decomposition can be specifically tailored to a particular class of problems.  
The inspiration for such decompositions arises out of matrix factorizations such the eigenvalue decomposition (EVD), the singular value decomposition (SVD), and nonnegative matrix factorization (NMF).  For example, the SVD factorizes matrix $\textbf{A} \in \mathbb{R}^{m \times n}$ into a sum of rank-1 matrices,
\begin{equation}
    \textbf{A} = \textbf{USV}^{\mathsf{T}} = \sum_{i=1}^{\min \{m,n\}} s_i \cdot \textbf{u}_i \textbf{v}_i^{\mathsf{T}}
\end{equation}
where $\textbf{U} \in \mathbb{R}^{m\times m}, \textbf{V} \in \mathbb{R}^{n \times n}$ are orthogonal matrices and $\textbf{S} \in \mathbb{R}^{m \times n}$ is a nonnegative diagonal matrix where the values are ordered from highest to lowest.  The SVD can be written in terms of $n$-mode products as,
\begin{equation}
    \textbf{A} = \textbf{S} \times_1 \textbf{U} \times_2 \textbf{V}
\end{equation}
Generalization of different EVD/SVD properties lead to multitudes of tensor decompositions \cite{lathauwerMatrixTensorMultilinear1997}.

The Tucker decomposition and the canonical polyadic (CP) are two tensor decompositions that are higher-order analogies to the SVD.  
The Tucker decomposition factorizes a tensor into a series of factor matrices and a core tensor.  
When these factor matrices are constructed via SVD on the modal unfoldings of a tensor, the Tucker decomposition is called the HOSVD \cite{delathauwerMultilinearSingularValue2000}.  
This approach is well suited for compression purposes and to capture concise representative models. 
The Tucker decomposition uses the notion of multilinear rank as a measure of tensor rank, e.g. multilinear rank(L,M,N) for a mode-3 tensor.

\begin{figure}[h]
  \centering
  \includegraphics[height=1.5in]{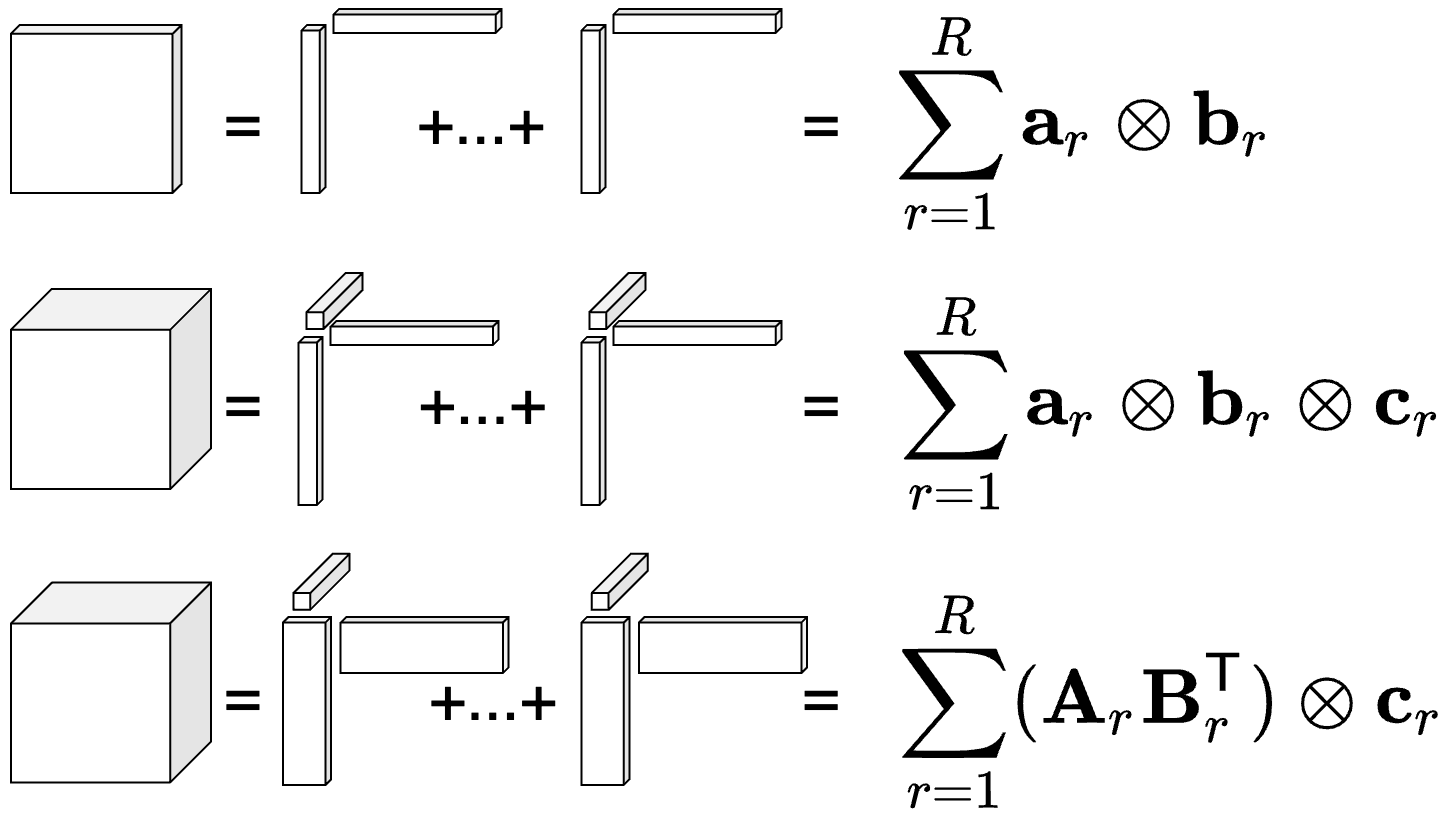}
  \caption{Matrix factorization versus tensor decomposition, Canonical Polyadic (middle), multilinear rank ($L_r, L_r, 1$) decomposition (bottom).}
  \label{sum_rank1_tensor}
\end{figure}

The CP decomposition (Figure \ref{sum_rank1_tensor}-middle) factorizes a tensor, $\mathbf{\mathcal{X}} \in \mathbb{R}^{I_1 \times I_2 \times \cdots \times I_d}$, as a linear combination of $R$ rank-1 tensors in the form  
\begin{equation*}
    \boldsymbol{\mathcal{X}} \approx \boldsymbol{\mathcal{M}} = \sum_{r=1}^{R}\lambda_r\textbf{a}_r^{(1)}\circ \textbf{a}_r^{(2)}\circ\dots\circ \textbf{a}_r^{(d)},
\end{equation*}
where $\lambda_r$ is a scalar weight, $\textbf{a}_r^{(k)}$ is a column vector $\in \mathbb{R}^{I_k}$, and $\circ$ is the outer product.  
The $k$-th mode factor matrix is formed from the respective column vectors of each rank-1 term the summation as $\textbf{A}^{(k)} = [\textbf{a}_1^{(k)} \cdots \textbf{a}_R^{(k)}] \in \mathbb{R}^{I_k \times R}$.  
A CP model tensor is sometimes referred to as a Kruskal tensor (K-tensor) and denoted as $\boldsymbol{\mathcal{M}} = \llbracket\boldsymbol{\lambda};\textbf{A}^{(1)},\textbf{A}^{(2)},\dots,\textbf{A}^{(d)}\rrbracket$ where $\boldsymbol{\lambda} = [ \lambda_1 \cdots \lambda_R]^{\mathsf{T}}$ is a weight vector and $\{ \textbf{A}^{(1)}, \dots, \textbf{A}^{(d)}\}$ are the factor matrices.
Note, absence of $\boldsymbol{\lambda}$ implies the weight vector has been absorbed into the factor matrices.

The common approach to computing solutions to CP decompositions is by solving the following nonlinear least-squares problem,
\begin{equation}\label{eq:CP_problem_def}
    \min_{\boldsymbol{\mathcal{M}}}\|\boldsymbol{\mathcal{X}}-\boldsymbol{\mathcal{M}}\|^2 \quad \text{s.t} \quad \boldsymbol{\mathcal{M}} = \llbracket\boldsymbol{\lambda};\textbf{A}^{(1)},\textbf{A}^{(2)},\dots,\textbf{A}^{(d)}\rrbracket
\end{equation}
where the minimization is in regards to the weight vector and factor matrices.  
There have been a wide variety of approaches developed for solving Eq. \ref{eq:CP_problem_def}, but the most commonly successful approach for computing a solution is an iterative alternating least-squares (ALS) method that cycles through the modes $k=1,\dots,d$ with each iteration, holding the modes other than $k$ fixed and solves the resulting linear least square for $\textbf{A}^{(k)}$. An alternate approach to ALS methods are gradient based methods such as damped Gauss-Newton and its variant PMF3, included in Tomasi and Bro's survey of CP algorithms\cite{tomasiComparisonAlgorithmsFitting2006}. 

The block-term decomposition (BTD) \cite{delathauwerDecompositionsHigherOrderTensor2008a} decomposes a tensor into a sum of multilinear rank terms, i.e. a sum of tucker decompositions, and is a generalization that captures both the CP and Tucker decompositions as special cases.
For a given tensor, $\boldsymbol{\mathcal{X}} \in \mathbb{K}^{I_1 \times I_2 \times \cdots \times I_d}$, the BTD is given as
\begin{equation}
    \boldsymbol{\mathcal{X}} \approx \boldsymbol{\mathcal{M}} = \sum_{r=1}^R \boldsymbol{\mathcal{D}}_r \times_1 \textbf{A}^{(1)}_r \times_2 \textbf{A}^{(2)}_r \times_3 \cdots \times_d \textbf{A}^{(d)}_r,
\end{equation}
where $\boldsymbol{\mathcal{D}}_r \in \mathbb{R}^{L_1 \times L_2\times \cdots \times L_d}$ is a core tensor of full multilinear rank ($L_{1_r},L_{2_r},\dots,L_{d_r}$) and $\textbf{A}^{(k)}_r \in \mathbb{K}^{I_k \times L_{k_r}}$ (with $I_k \geq L_{k_r}$) is a factor matrix for the $k$-th mode of full column rank, for all $1 \leq k \leq d$ and $1 \leq r \leq R$.

Another special case the BTD captures approximates a tensor as a sum of multilinear rank ($L_r , L_r, 1)$ terms (\ref{sum_rank1_tensor} - bottom) that has been shown to be effective across a wide range of applications from signal processing to compression of neural network (NN) models \cite{domanovUniquenessComputationDecomposition2020}. 
The tensor decompositions captured by BTD are of particular interest as they provide unique factorizations under relatively mild conditions \cite{bhaskaraUniquenessTensorDecompositions2013, domanovUniquenessComputationDecomposition2020, zhouEfficientNonnegativeTucker2015}. 
A unique decomposition is desirable in the context of DC as it can be treated as an explanatory model of the data that can be more easily interpreted by a domain expert. 

Tensor decomposition as a tool for low rank approximation has vast potential for facilitating DC for large scale data.
Udell and Townsend \cite{udellWhyAreBig2019} provide theoretical results that indicate big data is inherently low-rank, at least in the matrix case.
They show that the rank $r$ of lower-rank matrix $\textbf{X}$ that approximates a matrix $\textbf{A} \in \mathbb{R}^{m \times n}$ grows as $\mathcal{O}(\log (m + n))/ \epsilon^2$.
While this has yet to be demonstrated for data tensors, there are reasons to believe it is likely \cite{udellBigDataLow2019}.

\subsubsection{Computational Challenges Associated With Tensors} \label{sec:Tensors:Challenges}
There are several computational challenges associated with tensor-based methods, many of which arise from the so-called 'curse of dimensionalty' where the number of elements of a tensor scales exponentially as the order of a tensor increases.
This often leads to sparsity in the data, memory management and computational complexity issues, ill-conditioning, complicated optimization landscapes with increased numbers of local minima, and scalability challenges.
An additional challenge posed by tensors is the fact that most problems relating to tensors are known to be NP-hard  or NP-complete \cite{hillarMostTensorProblems2013}.

\subsection{Existing Tensor Survey Review}
\label{sec:Tensors:ExistingSurveys}
The following review of surveys targeting tensors and TDA is presented to illuminate the  spectrum of problem domains where TDA methods are being deployed.  The intent of this review is to provide the reader with a better sense of how TDA methods can be interwoven to suit a given application (e.g. the use of tensor decomposition models in many tensor completion and tensor clustering methods) and is not meant to be exhaustive.  To the best of our knowledge, there are currently no surveys of TDA as applied specifically in the space of DC.

\subsubsection{Data Mining and Analytics}
In the space of data mining and analytics, Acar and Yener \cite{acarUnsupervisedMultiwayData2009} explore three groupings of tensor decomposition models (CP-family, Tucker-family, and alternative models) with applications in social network analysis, text mining, and computer vision.
Fanee-T and Gama review tensor-based anomaly detection in terms of learning techniques, TD, and tensor rank estimation \cite{fanaee-tTensorbasedAnomalyDetection2016}.
Song et al. \cite{songTensorCompletionAlgorithms2019} reviews tensor completion algorithms in big data analytics across three families of completion algorithm: Trace-norm based, Decomposition based (CP and Tucker), and other variants (e.g. Non-negative constrained approaches).
Thanh et al. \cite{thanhContemporaryComprehensiveSurvey2023} reviews advances in online TDA with coverage of five tensor decompositions: CP, Tucker, BTD, TT, and t-SVD.  Special attention is given streaming CP decomposition algorithms categorized by approach: subspace-based, block-coordinate descent based, Bayesian inference band, and multi-aspect streaming models.
Papalexakis et al \cite{papalexakisTensorsDataMining2016} survey tensors and tensor factorizations as tools in data mining and data fusion with an emphasis on scalability of algorithms.
These works illuminate the tensor's versatility in the space of large-scale data mining and its suitability for modern workloads, demonstrating how methods for TDA can be adapted to a wide variety of data environments, e.g. missing data, nonnegative data, streaming data, with deliberate problem formulation and choice of measure such as rank type or norm type.
In addition, these works highlight how tensor decomposition leverages seperability to explore the ways complex datasets can be expressed in terms of simpler components to improve insights and the efficiency of computations.

\subsubsection{Machine Learning and Neural Networks}
Tensors and TDA appear in multiple contexts across a variety of ML applications.
Lu et al. \cite{luSurveyMultilinearSubspace2011} covers supervised and unsupervised multilinear subspace learning (MSL) by reviewing higher-order extensions of principle component analysis (PCA) and linear discriminant analysis (LDA), two approaches in linear subspace learning.
MSL algorithms are grouped according to learning type (supervised or unsupervised), then by projection type (tensor-to-tensor or tensor-to-vector), and lastly by the objective type the learning types (unsupervised/supervised) optimize for.
Sidiropoulos et al. \cite{sidiropoulosTensorDecompositionSignal2017} explores tensor decomposition models; CP, Tucker, multilinear singular value decomposition (MLSVD) and HOSV, with special attention given to rank estimation and uniqueness properties across ML applications of source separation, collaborative filtering, mixture and topic modeling, classification, and MSL.
Sun et al. \cite{sunTensorsModernStatistical2021} provides overview of tensor-based methods in statistical learning as organized by learning topic: tensor supervised learning (e.g. tensor regression), tensor unsupervised learning (e.g. tensor clustering), tensor reinforcement learning (e.g. stochastic low-rank tensor bandit or learning Markov decision process via tensor decomposition), and tensor deep learning (e.g. tensor-based network compression or deep learning theory through tensor methods).  
Chao et al. \cite{chaoSurveyMultiviewClustering2021} review multi-view clustering (MVC) which extends standard clustering to tensor datasets and apply a ML-centric taxonomy to categorize MVC algorithms as generative or discriminative (which can be further split into five groups: Common eigenvector matrix, common coefficient matrix, common indicator matrix, direct combination, and combination after projection.).
Chen et al. \cite{chenRepresentationLearningMultiview2022} cover MVC based in terms of representation learning and non-represention learning models. Coverage of representation learning models is split into shallow representation learning and deep representation learning models.

Tensors and TDA are playing an increasing role in the space of deep learning and NNs owing in part to the fact that the basic building blocks neural architectures can be viewed as multilinear mappings represented by tensors. 
Panagakis et al. \cite{panagakisTensorMethodsComputer2021} review TDA in terms of representation learning and DNN architecture through the lens of computer vision and visual data tensors.  The authors accompany the survey with a repository of tutorials for each of the methods and architectures covered.
Wang et al. \cite{wangTensorNetworksMeet2023} survey tensorial neural networks (TNNs), architectures that combine TNs and NNs, with attention given to network compression, information fusion, and quantum circuit simulation.
Liu and Parhi \cite{liuTensorDecompositionModel2023a} review the use of different tensor decomposition to replace network layers with low-rank tensor approximations.  
The authors focus on NNs in computer vision and natural language processing with an emphasis on compressing the parameterization of convolutional NNs (CP, Tucker, and Tensor Train decompositions), recurrent NNs/LSTMs (Tensor Train, Tensor Ring, Block-Term, and Hierarchical Tucker decompositions) , and Transformers (Tensor Train and Block-Term decompositions).

These works demonstrate how tensors tensor methods can be readily used to extend ML applications to higher order datasets through creative problem formulation.
In terms of DC, tensor-based implementations of ML models opens the door for ML model-based DC akin to landmarking meta-features or model-based meta-features (e.g. depth of a tensor decision tree).

\subsubsection{Signal Processing}
Tensors and TDA fit naturally into the space signal processing owing to the tensor's ability to represent signals in multiple dimensions (e.g. a color video is a mode-4 tensor with 2 spatial modes, a temporal mode, and a color mode) and have been utilized since the inception of the field of study. 
Muti and Bourennane \cite{mutiSurveyTensorSignal2007} review tensor-based algebraic filtering methods using higher-order extensions of subspace projection (e.g. PCA) or higher-order mean squared error minimization (e.g. higher-order Wiener filters).
Cichocki et al. \cite{cichockiTensorDecompositionsSignal2015a} provide a comprehensive survey of tensor decomposition in the context of signal processing with attention to the multilinear algebra that enables tensor decomposition to naturally generalize commonly used signal processing paradigms like canonical correlation, linear regression, signal separation, feature extraction, and classification.
Cong et al. \cite{congTensorDecompositionEEG2015} survey how CP and Tucker decompositions are used in electroencephalograpy (EEG) signal processing with attention given to rank estimation and higher-order methods in partial least squares.
The application of tensors to the problem space of signal processing demonstrates how the principles of separability and multilinearity embodied by tensors can be leveraged to coax improved insights and understanding from datasets.

\subsubsection{Numerical Multilinear Algebra }
Numerical MLA extends the principles of linear algebra to higher-order settings and is the backbone that make tensors and TDA so broadly applicable.

Qi et al. \cite{qiNumericalMultilinearAlgebra2007} provide a survey of topics in numerical MLA  that includes: tensor decomposition, rank estimation, low-rank approximation, and numerical stability.
Chang et al. \cite{changSurveySpectralTheory2013} survey developments in the spectral theory of nonnegative tensors with attention given to applications in higher-order Markov chains, spectral theory of hypergraphs, and quantum entanglement.
Lim \cite{limTensorsComputations2021} reviews of the theory underlying the general utility of tensors in context to three mathematical principles that arise across computation and are embodied through tensors by definition: equivariance (e.g. transformation-based approach like Fourier transforms or a sequences of Givens rotations), separability (e.g. tensor product construction of various objects such as bases, function spaces, operators, and kernels), and multilinearity.
These works present provide a sense of where the theoretical understanding of tensors currently stands.
The boundary between the theory of tensors and implementation (i.e. numerical multilinear algebra) ultimately dictates what is possible in terms of tensor-based algorithms.

\subsubsection{Hardware, Software, and Scalability}
The general utility of tensors in computational application has lead to research and development at the implementation level.
Adaptation of tensors and TDA has resulted in a growing multitude of software packages, Psarras et al. \cite{psarrasLandscapeSoftwareTensor2021} surveys 77 packages in terms of functionality in the following categories: data manipulation, element-wise operations, general contractions, specific contractions, tensor decomposition, code language, tensor type (e.g. dense, sparse, symmetric), and target system (e.g. CPU, GPU, distributed memory).
Mass incorporation of TDA has also led to demand for scalable algorithms to accommodate the massive, often sparse, datasets of big data.
Papalexakis et al. \cite{papalexakisLargeScaleTensor2013} discuss the challenges in scaling tensor decomposition to big data and cover two successful means that achieve scalability via simplification of costly operations coupled with use of the Map/Reduce environment and a tensor sketch approach with biased sampling, respectively.

In addition to software and algorithmic design, research has been directed at the development of hardware and hardware-accelerators for tensor computations.
Dave et al. \cite{daveHardwareAccelerationSparse2021} reviews efforts at accelerating  sparse tensor calculations through the use of parallel platforms and Xiao et al. \cite{xiaoSurveyAcceleratingParallel2023} surveys recent advancements hardware acceleration for tensor computations in ML models.
Menghani \cite{menghaniEfficientDeepLearning2023} reviews advancements in the pursuit of efficient deep learning including discussion of NVIDIA's use of Tensor Cores in GPUs \cite{VoltaWorldMost2017} and Google' use of systolic arrays in its TPU \cite{jouppiTenLessonsThree2021, kangWhySystolicArchitectures1982} to provide an efficient hardware implimentation of expensive matrix multiplication operation at the heart of matricized formulations of TDA methods, see Eq. \ref{eq:matricized mode-n product}.

These works illustrate how tensors and TDA are increasingly supported at the implementation level through the development of numerical MLA libraries and domain-specific hardware design.

\subsection{Dataset Characterization with Tensors and TDA} \label{sec:Tensors:Tensor-based Characterization}
To the best of our knowledge there are limited works in the tensor literature that specifically target the DC problem, nonetheless there are recent works and momentum conducive to developing a framework for tensor-based DC.  

The following works are grouped according the types of dataset characteristics used in the comparative analysis workflow from Subsection \ref{sec:Dataset Characterization:ComparativeAnalysis} (i.e. statistical, structural, and model-based characteristics).

\subsubsection{Statistical Characterization}

The Generalized Canonical Polyadic tensor decomposition (GCP) \cite{hongGeneralizedCanonicalPolyadic2020} treats the solution of the CP decomposition as a maximum likelihood estimate (MLE) so as to afford for a multitude of statistically motivated loss functions.  As a consequence of this modification, the GCP becomes an unsupervised learning method that approximates the natural parametrization of the distribution assumed to be associated with a tensor data instead of the data itself.
This imbues the GCP with an ability to act as an exploratory tool for testing statistical hypotheses regarding a data tensor, e.g. the loss function(s) that yield the lowest CP-rank decomposition or model with best fit corresponds to statistical assumption(s) about a dataset.
The GCP is covered in further detail in Subsection \ref{sec:GCP} as an illustrative example how tensor methods are malleable to specific dataset considerations.

Additionally, several tensor-based methods for nonparametric probability density estimation have been recently proposed.
The flexibility afforded by nonparametric approaches come with an increased computational complexity as they rely on the calculation over many data points.
Mixture models in statistics are probabilistic models that assume a dataset is generated from a combination of distributions associated with specific subgroups contained that allow for improved modeling of heterogeneity in a dataset.  Tensors fit into mixture models as structures for model parametrization, e.g. the $d$-th moment of a $n$-dimensional random variable is a mode-$d$ symmetric tensor of size $n^d$.

\cite{pereiraTensorMomentsGaussian2022} introduced theory and efficient computational strategies for parameter estimation in Gaussian mixture models (GMMs) via tensor moments which was built upon by \cite{zhangMomentEstimationNonparametric2022} to develop a method for the nonparametric estimation of conditionally-independent mixture models in $\mathbb{R}^n$ formulated as,
\begin{equation}\label{eq:Mixture Models}
    \mathcal{D} = \sum_{j=1}^r w_j \bigotimes_{i=1}^n \mathcal{D}_{ij},
\end{equation}
where $r$ is the number of mixture components with each mixture component is a product of $n$ distributions,
and the vector $\textbf{w} = (w_1,\dots,w_r)$ is the mixture weights for the model.
Both works rely upon implicit symmetric tensor decompositions in the calculation of higher-order tensor moments (i.e the aforementioned $d$-order symmetric tensors),
\begin{equation} \label{eq:d-th moment tensor}
    \frac{1}{p} \sum_{i=1}^p \textbf{v}_i^{\otimes d} \approx \mathbb{E}_{X \sim \mathcal{D}} \left[ X^{\otimes d}\right].
\end{equation}
These works leverage implicit methods \cite{shermanEstimatingHigherOrderMoments2020} to circumvent the computational and storage costs associated with using tensor moments.

Wu et al. \cite{wuTERMModelTensor2023} present an alternate method for tensor-based nonparametric density estimation through a tensor ring mixture model, TERM, that is inspired by ensemble learning and the distinct information provided by lower weight components in GMMs.
TERM relies upon the tensor ring decomposition \cite{zhaoTensorRingDecomposition2016} and its unique rotational invariance via a tensor ring density estimator (TRDE) that approximates coefficients of the expansion of the density function in some uniform B-spline basis via tensor ring factorization.
The TRDE allows exact sampling as well as the calculation of the cumulative density function, conditional probability density function and precise computation of the partition function.

These works present methods that could significantly enhance the statistical characterization of datasets, especially in complex, high-dimensional, and heterogeneous data environments.  Tensor moments bring an added layer of depth to understanding and modeling the distribution of multidimensional data. Whether in parametric or nonparametric probability density estimation, they offer a means to quantify and leverage the latent relationships of the data in multiple dimensions.

\subsubsection{Structural Characterization} \label{sec:Tensors:Structural-based Characterization}
One of the many appeals of tensors is their ability to robustly preserve the latent spaces and structures formed by multi-relational datasets.
Domanov and De Lathauwer \cite{domanovComputationComparisonTensor2021} leverage this ability to present a novel method for the comparison of two tensors, $\boldsymbol{\mathcal{A}}, \boldsymbol{\mathcal{B}}$, that treats the tensor decomposition as pattern recognition.
This work shows that column spaces of modal unfoldings of the tensors, namely
\begin{equation}
    \text{col}(\textbf{B}_{(S^c;S)}) \subseteq \text{col}(\textbf{A}_{(S^c;S)}), \quad n \in \{1,\dots,N\},
\end{equation}
where $\textbf{A}_{(n^c;n)}$,$\textbf{B}_{(n^c;n)} \in \mathbb{R}^{\frac{I_1\cdots I_d}{I_n} \times I_n}$,
is a sufficient condition to show  $\boldsymbol{\mathcal{B}}$ is generated by terms (possibly scaled) from the decomposition of $\boldsymbol{\mathcal{A}}$.
The number of terms and their associated multilinear ranks can be identified by this method as well.  
The process is framed in terms of linear algebra which allows the method to \emph{implicitly} address tensor decomposition (CP decomposition or BTD) to bypass computational difficulties like ill-conditioning and NP-hardness higher order formulations incur.
An addition aspect of this work of interest in terms of DC is the use of tensorization to modify the problem space of blind source separation (BSS) to accommodate a tensor-based approach.
This embodies malleable formulation in a different direction from the GCP example (i.e. the dataset is tailored to the tools of TDA).
See Subsection \ref{sec:Tensorization} for an in depth review as an illustrative example of properties of tensors and TDA that can be leveraged to provide improved characterization of datasets: malleable formulation and implicit formulation.

Tensor clustering generalizes matrix-based clustering to tensor datasets with the aim of preserving the latent structure of multi-dimensional data that is lost when flattened into matrices or vectors.
Clustering is given as a method for the structural characterization of data in Subsection \ref{sec:Dataset Characterization:ComparativeAnalysis}.
Jegelka et al. \cite{jegelkaApproximationAlgorithmsTensor2009} claim to be the first to present an approximation algorithm for tensor clustering that admits a broad class of objective functions.
This has lead to the development of a variety of methods for tensor-based clustering frameworks including: multi-view clustering \cite{chaoSurveyMultiviewClustering2021, chenRepresentationLearningMultiview2022, fangComprehensiveSurveyMultiView2023a}, co-clustering \cite{biernackiSurveyModelBasedCoClustering2023}, and tensor spectral clustering \cite{bensonTensorSpectralClustering2015, jiaMultiViewSpectralClustering2021, wuGeneralTensorSpectral2016}.

Chi et al. \cite{chiProvableConvexCoclustering2020} remedy non-convexity in tensor clustering with their convex co-clustering (CoCo) procedure that provides a strongly convex formulation for the problem of co-clustering a $D$-way array for $D \geq 3$ and is another example of malleability in tensor-based methods.  
The CoCo estimator reveals a "blessing of dimensionality" that does not exist in convex formulations of vector or matrix based clustering where it remains consistent, in terms of statistical guarantee, as the order of a tensor increases even if the number of underlying co-clusters grows as a function of the number of elements of a tensor.  
For a observed mode-3 tensor, $\boldsymbol{\mathcal{X}} \in \mathbb{R}^{I_1 \times I_2 \times I_3}$, the model is given as a sum of means tensor $\boldsymbol{\mathcal{U}}^{\ast}$ whose elements are expanded from a co-cluster means tensor $\boldsymbol{\mathcal{C}}^{\ast} \in \mathbb{R}^{K_1 \times K_2 \times K_3}$,
\begin{equation}
    \boldsymbol{\mathcal{U}}^{\ast} = 
    \boldsymbol{\mathcal{C}}^{\ast} \times_1 
    \textbf{M}_1 \times_2
    \textbf{M}_2 \times_3
    \textbf{M}_3, 
\end{equation}
where $\textbf{M}_d \in \{0,1\}^{I_d \times K_d}$ is the membership matrix for the $d$th mode where the $ik$th element of $\textbf{M}_d$ is one if and only if the $i$th mode-$d$ slice belongs to the $k$th mode-d cluster for $k \in \{1,\dots,K_d\}$.
The CoCo formulation draws inspiration from the successes of Lasso \cite{tibshiraniRegressionShrinkageSelection1996} to simultaneously identify partitions along the modes of $\boldsymbol{\mathcal{X}}$ to estimate $\boldsymbol{\mathcal{C}}^{\ast}$ through the minimization of the convex objective function 
\begin{equation} \label{eq:CoCo Objective}
    F_{\gamma}(\boldsymbol{\mathcal{U}}) =
    \frac{1}{2} \|\boldsymbol{\mathcal{X}} -
    \boldsymbol{\mathcal{U}}\|_F^2 +
    \gamma R(\boldsymbol{\mathcal{U}}),
\end{equation}
where $R(\boldsymbol{\mathcal{U}}) = R_1(\boldsymbol{\mathcal{U}}) + R_2(\boldsymbol{\mathcal{U}}) + R_3(\boldsymbol{\mathcal{U}})$ and
\begin{equation}
    \begin{split}
        R_1(\boldsymbol{\mathcal{U}}) &= \sum_{i < j} w_{1,ij} \| \boldsymbol{\mathcal{U}}_{i::} -
        \boldsymbol{\mathcal{U}}_{j::} \|_{F} \\
        R_2(\boldsymbol{\mathcal{U}}) &= \sum_{i < j} w_{2,ij} \| \boldsymbol{\mathcal{U}}_{:i:} -
        \boldsymbol{\mathcal{U}}_{:j:} \|_{F} \\
        R_3(\boldsymbol{\mathcal{U}}) &= \sum_{i < j} w_{3,ij} \| \boldsymbol{\mathcal{U}}_{::i} -
        \boldsymbol{\mathcal{U}}_{::j} \|_{F}.
    \end{split}
\end{equation}
By seeking the minimizer $\boldsymbol{\mathcal{\hat{U}}} \in \mathbb{R}^{I_1 \times I_2 \times I_3}$, co-custering is cast as a signal approximation problem modeled as a penalized regression where the quadratic term in Eq. \ref{eq:CoCo Objective} quantifies how well $\boldsymbol{\mathcal{U}}$ approximates $\boldsymbol{\mathcal{X}}$, the regularization term $R(\boldsymbol{\mathcal{U}})$ penalizes deviations from a checkerboard pattern, and $\gamma$ parameterizes the balance between the two. The weights, $w_{d,ij}$ quantify the similarity of the $i$th and $j$th mode-$d$ slices.

He et al. \cite{heDetectingNumberClusters2010} apply tensor methods as means of determining the number clusters in $n$-way probabilistic clustering and draw a theoretical connection between the number of components in a CP model (i.e. the CP rank) and the cluster number.

Consequently, determining the rank of a tensor presents another opportunity for the structural characterization of datasets.
Although determining the rank of a tensor is NP-hard, the success of low rank approximations is driven by an ability to identify a model with the a sufficient number of components to meaningfully model the data.
Matrix amplification is a method for identifying an ideal number of components for a low-rank approximation.
Given a rank-$r$ matrix \textbf{A}, the singular values are amplified by taking advantage of the orthogonality of the factor matrices \textbf{U},\textbf{V} from the SVD of \textbf{A} to form the matrix,
\begin{equation}
    \textbf{AA}^T\textbf{A} =
    (\textbf{U}\boldsymbol{\Sigma}\textbf{V}^T)
    (\textbf{V}\boldsymbol{\Sigma}^T\textbf{U}^T)
    (\textbf{U}\boldsymbol{\Sigma}\textbf{V}^T) =
    \textbf{U}\boldsymbol{\Sigma}^3 \textbf{V}^T.
\end{equation}
The matrix $\textbf{AA}^T\textbf{A}$ the singular values $\lambda_1^3 > \lambda_2^3 > \dots > \lambda_r^3$ which increases the ratio between consecutive singular values from $\lambda_i / \lambda_{i+1}$ to $\lambda_i^3 / \lambda_{i+1}^3$. 
This is useful in terms of low-rank approximation where the ratio between successive singular values provides a threshold measure for determining an appropriate number of singular values $<r$ for the approximation. 
Tokcan et al. \cite{tokcanAlgebraicMethodsTensor2021} presents an algorithm that generalizes the notion of matrix amplification to tensors through the introduction of colored Brauer diagrams (a modification of a concept from classical invariant theory) used for the description of tensor invariants of the orthogonal group in order to define approximations of the spectral and nuclear norms as well as achieve and ALS-style algorithm for tensor amplification.

In the same way that matrices are a natural representation of a graph, adjacency tensors are a means of representing hypergraphs or multigraphs.  
Recent works have demonstrated the usefulness of tensor methods applied to cases dealing with underlying hypergraph or multigraph data and clustering.  
Papalexis et al. \cite{papalexakisMoreViewsGraph2013} present a multi-graph clustering technique that employs a version of the CP decomposition with sparsity constraints imposed on the factor matrices.  
More recently, Ouvrard et al. \cite{ouvrardAdjacencyTensorRepresentation2018} provide a definition for adjacency that allows for the uniformization of a non-uniform hypergraph resulting in an $e$-adjacency tensor that preserves both the structural information and the information on each of the hyperedges in the hypergraph.  
This formalism may facilitate the efficacy of tensor methods and provide additional means for the structural characteristization of data tensors. 

\subsubsection{Model-based Characterization} \label{sec:Tensors:Model-based Characterization}
Tensors are increasingly present in a variety ML applications \cite{jiSurveyTensorTechniques2019} in a multitude of ways.  
There are two tensor-based approaches for ML that are of interest in terms of model-based DC.  

The first approach consists of methods that modify existing ML algorithms to accommodate tensor data and include the likes of support tensor machines \cite{biswasLinearSupportTensor2017,haoLinearSupportHigherOrder2013}, tensor decision trees \cite{krawczykTensorDecisionTrees2021}, generative models \cite{shiaoTenGANAdversariallyGenerating2024} and tensor classification \cite{maruhashiLearningMultiWayRelations2018, phanTensorDecompositionsFeature2010}.
Methods of this sort provide a direct road map for model-based DC akin to landmark-based meta-features (i.e. performance of 'simple' ML models) and model-structure meta-features (e.g. depth of decision tree or number of internal nodes) discussed in \ref{sec:Meta-Features} or model performance characteristics (e.g. classifier accuracy) detailed in \ref{perfom_comp}.

The second approach is the incorporation of tensors at the architectural level in NNs.
Tensors in NN architectures can occur at the layer level \cite{kossaifiTensorRegressionNetworks2017,maTensorizedTransformerLanguage2019a} up to an entire network parameterized by a single tensor \cite{kossaifiTNetParametrizingFully2019}.
Newman et al. \cite{newmanStableTensorNeural2018} present \emph{tensor neural networks} ($t$-NNs) based on the $t$-product \cite{kilmerFactorizationStrategiesThirdorder2011}, an algebraic formulation for multiplying tensors based on circulant convolution.  
The $t$-NN framework recasts traditional fully connected layers formulated as,
\begin{equation}
    \textbf{A}_{j+1} = \sigma (\textbf{W}_j \cdot \textbf{A}_j + \textbf{b}_j )
\end{equation}
in terms of tensors and tensor operations,
\begin{equation}
    \boldsymbol{\mathcal{A}}_{j+1} = \sigma ( \boldsymbol{\mathcal{W}}_j \ast \boldsymbol{\mathcal{A}}_j + \boldsymbol{\mathcal{B}}_j )
\end{equation}
where $\boldsymbol{\mathcal{A}}_{j+1},\boldsymbol{\mathcal{A}}_j,\boldsymbol{\mathcal{W}}_j$ are tensors and $\ast$ is the $t$-product as defined in \cite{kilmerFactorizationStrategiesThirdorder2011}.  
The $t$-product possesses a matrix-mimetic structure that allows deep NNs to be interpreted as discretizations of non-linear differential equations which promotes superior generalization.
Given $\boldsymbol{\mathcal{A}} \in \mathbb{R}^{l \times p \times n}$ and $\boldsymbol{\mathcal{B}} \in \mathbb{R}^{p \times m \times n}$ the $t$-product is 
\begin{equation}
    \boldsymbol{\mathcal{C}} = 
    \boldsymbol{\mathcal{A}} \ast
    \boldsymbol{\mathcal{B}} = 
    \fold \left(\bcirc(\boldsymbol{\mathcal{A}})\cdot \unfold(\boldsymbol{\mathcal{B}})\right)
\end{equation}
where $\boldsymbol{\mathcal{C}} \in \mathbb{R}^{l \times m \times n}$ and $\bcirc(\cdot)$ constructs a block circulant matrix using the frontal slices of $\boldsymbol{\mathcal{A}}$.
The block circulant structure of the $t$-product allows the operation to be implemented as independent matrix multiplications in the Fourier domain using fast Fourier transform on the frontal slices of $\boldsymbol{\mathcal{A}}$ and $\boldsymbol{\mathcal{B}}$ which affords a perfectly parallelizable algorithm.
This work captures the power of tensors as an abstraction that efficiently leverages natural structuring of data coupled with algebraic certainty.
Additionally, the $t$-product embodies the malleability of tensor methods where the tensor-tensor product operator is defined in a computationally advantageous manner.

The second approach gives rise to a potential for calculating improved model-structure characterizations    through the analysis of the weight tensors of model learned on a dataset and is left as an open question.

One such implication is the potential ability to observe the network mechanics during training and inference with tensor based NN architectures thereby contributing an additional view classification characterization of data in addition to the type of performative analysis detailed in \ref{perfom_comp}.

\section{Properties of Tensors Conducive to Improved Dataset Characterization: Illustrative Examples} \label{sec:Illustrative Examples}
The GCP \cite{hongGeneralizedCanonicalPolyadic2020} and a novel method for tensor comparison \cite{domanovComputationComparisonTensor2021} are presented as illustrative examples of two properties of tensors and tensor methods that are powerful in terms of potential for improved DC: malleable problem formulations and implicit solutions. 

\subsection{The Generalized Canonical Polyadic Decomposition: Statistical Characteristic Informed Tensor Factorization}
\label{sec:GCP}
The CPD is well suited for DC as it possesses: a relativity simple formulation (e.g. sum of rank-1 terms) that fosters ease of interpretability, a measure of structure in the form of rank, uniqueness guarantees under mild conditions, a malleable formulation that can be tailored to the specifics of a problem space (e.g. count or nonnegative data), and software libraries spanning a multitude of platforms and target architectures \cite{psarrasLandscapeSoftwareTensor2021}.  
These properties speak to the utility the CPD demonstrates across novel applications. For example, Hutter and Solomonik \cite{hutterApplicationPerformanceModeling2023} use a CP-model based tensor completion method to model and approximate application performance for various system configurations where tensor elements are execution times indexed according to a discretization of the input and configuration domains.  

The solution to the CP problem can be formulated according to the details of a dataset under consideration.
Chi and Kolda's \cite{chiTensorsSparsityNonnegative2012} alternating Poisson regression based CP solution and Ranadive and Baskaran's \cite{ranadiveAllOnceCP2021} all-at-once CP solution based on a generalized damped Gauss-Newton method are formulations of the CP problem that impose the assumption of a Poisson distribution on the data contained in a tensor (i.e. count data) and provide example of how the CP problem can be modified to specific details regarding a dataset under consideration.
The generalized canonical polyadic decomposition (GCP) is a recent formulation of CP problem that allows the loss function to be selected according to the nature of the dataset with the motivation to improve CPDs of strongly non-Gaussian data, e.g. binary, count, or nonnegative. 
These examples illuminate the malleability of methods in TDA.  

\subsubsection{The Generalized Polyadic Decomposition}
The novelty of the GCP decomposition \cite{hongGeneralizedCanonicalPolyadic2020} lies in posing the solution to the CP problem, Eq. \ref{eq:CP_problem_def}, as a maximum likelihood estimate such that the model $\boldsymbol{\mathcal{M}}$ is a MLE across the entries in $\boldsymbol{\mathcal{X}}$.  
This is accomplished by assuming a parameterized probability density function (PDF) or probability mass function (PMF) that yields a likelihood for each entry in $\boldsymbol{\mathcal{X}}$
\begin{equation}
    x_i \sim p(x_i|\theta_i) \text{ where } l(\theta_i) = m_i.
\end{equation}
where $l(\dot)$ is an invertible link function that takes the natural parameter, $\theta_i$, and connects it to the corresponding model parameter, $m_i$, for the observed variable, $x_i$.

While a Gaussian assumption is appropriate in many scenarios, such an assumption is not suitable when a dataset is strongly non-Gaussian (e.g. with binary or count data). 
The GCP computes an approximate CP model through the maximization of the likelihood $p(x_i | \theta_i)$ for each tensor entry $x_i$ via a link function $l(\theta_i) = m_i$ that connects the CP model parameter $m_i$ to the distribution parameter $\theta_i$. 
Link functions are a well known statistical concept with a body of work for their use in matrix factorizations\cite{gordonGeneralized2Linear2Models2002}.

In the general sense, the GCP model is an unsupervised learning method for parametric density estimation that approximates the natural parameter of the distribution underlying the assumed statistical model of the tensor data instead of modeling the tensor itself.
As such, the GCP provides a means of exploring a variety of statistical assumptions in decomposing a tensor so long as the assumed distribution has an invertible link function associated with it.

This yields the optimization problem,
\begin{equation}\label{eq:GCP_problem_def}
    \begin{split}
    \min_{\boldsymbol{\mathcal{M}}} &F(\boldsymbol{\mathcal{M};\boldsymbol{\mathcal{X}}}) = 
    \sum_{i \in \mathcal{I}}f(x_i,m_i) 
    \quad \\ 
    &\text{s.t.}\quad 
    \boldsymbol{\mathcal{M}} =
    \llbracket\textbf{A}^{(1)},\dots,\textbf{A}^{(d)}\rrbracket
    \end{split}
\end{equation}
where $f(x,m) = - \log p(x|l^{-1}(m))$ is referred to as the loss function.  

\subsubsection{Generalized Loss Functions}
The derivation of the Gaussian loss function, $f(x,m) = (x-m)^2$ \cite{hongGeneralizedCanonicalPolyadic2020}  begins with the common assumption that data is drawn from some latent low-rank structure and includes some noise such that,
\begin{equation}
    x_i = m_i + \epsilon_i \text{ with } \epsilon_i\sim\mathcal{N}(0,\sigma)\text{ for all } i\in\Omega
\end{equation}
where $\mathcal{N}(\mu,\sigma)$ is the Gaussian distribution with mean $\mu$ and standard deviation $\sigma$.  
By assuming $\sigma$ to be constant across $\boldsymbol{\mathcal{X}}$, the above can be rewritten 
$$
    x_i \sim \mathcal{N}(\mu_i,\sigma) \text{ with } \mu_i = m_i \text{ for all } i\in\Omega.
$$
Here, the link function is the identity, $m_i = l(\mu_i) = \mu_i$.  
Now considering the PDF for the normal distribution,
\begin{equation}
    p(x|\mu,\sigma) = \frac{e^{-(x-\mu)^2 / 2\sigma^2}}{\sqrt{2\pi\sigma^2}},
\end{equation}
it becomes the following when expressed as a negative log likelihood
\begin{equation}
    -\log p(x|\mu,\sigma) = \frac{(x-\mu)^2}{2\sigma^2} - \frac{1}{2}\log(2\pi\sigma^2).
\end{equation}
Substituting the model $m$ for the natural parameters, $\mu$ and $\sigma$, the element-wise loss function becomes
\begin{equation}
    f(x,m) = \frac{(x-m)^2}{2\sigma^2} - \frac{1}{2}\log(2\pi\sigma^2).
\end{equation}
The $\sigma$ terms are constant, thus have no impact on the optimization and can be removed, which yields the standard form
\begin{equation}
    f(x,m) = (x-m)^2
\end{equation}
rendering the Gaussian MLE as
\begin{equation}\label{eq:gcpProb}
    \min_{\textbf{A}^{(1)},\dots,\textbf{A}^{(d)}} F(\boldsymbol{\mathcal{M};\boldsymbol{\mathcal{X}}}) = \sum_{i=1}^{n^d}(x_i-m_i)^2.  
\end{equation}

Table \ref{tab:table1} presents a sample of potential loss functions for the GCP. 
The derivations and bounds for each loss function listed is covered in greater detail in \cite{hongGeneralizedCanonicalPolyadic2020}. 
Choice of appropriate loss function is a process that should be dictated by the form of the data.
Flexibility of loss function in GCP has the potential to be used for DC as a means of testing a variety of statistical assumptions in the analysis of a dataset. 

\begin{table*}[t]
    \begin{center}
        \caption{Examples of statistically motivated loss functions}
        \label{tab:table1}
        \begin{tabular}{|l|l|l|l|}
        \hline
        \textbf{Distribution}&\textbf{Link Function}&\textbf{Loss Function}&\textbf{Constraints}\\
        \hline \hline
        $\mathcal{N}(\mu,\sigma)$ & $m=\mu$ & $(x-m)^2$ & $x,m \in \mathbb{R}$ \\
        \hline
        {Bernoulli$(\rho)$} & $m =\rho/(1-\rho)$ & $\log(m+1) - x\log(m+\epsilon)$ & $x\in\{0,1\}, m\geq0$ \\
        & $m=\log(\rho/(1-\rho)$ & $\log(1+e^m)-xm$ & $x\in\{0,1\}, m \in \mathbb{R}$\\
        \hline
        Poisson$(\lambda)$ & $m=\lambda$ & $m-x\log(m+\epsilon)$ & $x\in\mathbb{N},m\geq0$\\
        & $m=\log\lambda$ & $e^m -xm$ & $x\in\mathbb{N}, m \in \mathbb{R}$\\
        \hline
        NegBinom$(r,\rho)$ & $m=\rho/(1-\rho)$ & $(r+x)\log(1+m)-x\log(m+\epsilon)$ & $x\in\mathbb{N}, m\geq0$\\
        \hline
        Gamma$(k,\sigma)$ & $m=k\sigma$ & $x/(m+\epsilon) + \log (m+\epsilon)$ & $x >0, m\geq 0$\\
        \hline
        Rayleigh$(\theta)$ & $m=\sqrt{\pi/2}\theta$ & $2\log(m+\epsilon) + (\pi/4)(x/(m+\epsilon))^2$ & $x>0,m\geq0$\\
        \hline
        \end{tabular}
    \end{center}
\end{table*}

\subsubsection{Reflections on the GCP}
The GCP serves as an two-fold example of the malleability of TDA methods through the reformulation of the CP problem as a MLE and the resulting gradient formulations used in computing a solution.
The flexibility to explore a variety of loss functions affords data analysts a novel approach for identifying the underlying statistical characteristics of a dataset through the use of tensor factorization. 
The relatively mild uniqueness guarantees afforded by CP decompositions provide a unique opportunity for comparing various statistical assumptions, but is encumbered by difficulty is terms of measuring similarities and the quality of CP models. 

\subsection{Comparison of Tensors: Leveraging Structural Characteristics of Data via TDA}
\label{sec:Tensorization}
Comparison of data is a fundamental application of DC.  
Domanov and De Lathauwer \cite{domanovComputationComparisonTensor2021} present a novel approach for the comparison of two tensors that relies heavily on the structural aspects of tensor decompositions.  
The authors exploit the algebraic structure of the problem to implicitly reframe higher-order calculations in terms of stable methods from numerical linear algebra which effectively circumvents many of the well known problems associated with tensors and tensor methods (i.e. NP-hardness of rank and ill conditioning).  
Additionally, the process of tensorization \cite{debalsTensorizationApplicationsBlind2017} is used to leverage assumptions about the underlying structural characteristics of a dataset to tailor the problem in terms of tensors and tensor methods. 

The method presented in \cite{domanovComputationComparisonTensor2021} is a departure from traditional approaches in that it treats tensor decomposition as pattern recognition by considering the "tensor similarity" problem:
\begin{itemize}
    \item How to verify that two $I_1 \times \cdots \times I_d$ tensors are generated by the same (possibly scaled) rank-1 terms?
    \item More generally, how to verify that two $I_1 \times \cdots \times I_d$ tensors are generated by the same (possibly scaled) ML rank-($L_1,L_2,\dots,L_d)$ terms?
\end{itemize}
Through this lens, Domanov and De Lathauwer lean on the seemingly trivial observation that if some tensor $\boldsymbol{\mathcal{B}}$ is a sum of (possibly scaled) terms from the decomposition of some tensor $\boldsymbol{\mathcal{A}}$, then $\text{col}(\textbf{B}_{(S^c;S)}) \subseteq \text{col}(\textbf{A}_{(S^c;S)})$ for all proper subsets $S$ of $\{1, \dots, d\}$ to forgo the explicit tensor computations.  
In fact, $d$ conditions are suffice to show $\boldsymbol{\mathcal{B}}$ is generated from the same (possibly scaled) terms as $\boldsymbol{\mathcal{A}}$ \cite[Section 4.2]{domanovComputationComparisonTensor2021}.  

The implications are that the matrices $\textbf{A}_{(n^c;n)}$ and $\textbf{B}_{(n^c;n)} \in \mathbb{R}^{\frac{I_1\cdots I_d}{I_n} \times I_n}$ can be used to compute the number of terms in the decompositions of $\boldsymbol{\mathcal{A}}$ and $\boldsymbol{\mathcal{B}}$ and the associated multilinear ranks in a manner that only relies on numerical linear algebra which can be reliably verified and implicitly computes the tensor decomposition.

\subsubsection{Matricizing the ML rank-($L_1,\dots,L_{\hat{d}},\cdot,\dots,\cdot$) decomposition}
Recall that the matrix $\textbf{A}_{(n^c;n)}$ is the matricization of the tensor $\boldsymbol{\mathcal{A}}$ where the column vectors consist of the vectorized mode-$n$ slices of the tensor.  
This allows the mode-$n$ product between a tensor $\boldsymbol{\mathcal{D}}$ and a matrix $\textbf{U}$ to be expressed in matrix form as
\begin{equation}\label{CCOT_eq:2.2}
    \boldsymbol{\mathcal{A}} = \boldsymbol{\mathcal{D}} \times_n \textbf{U}^{(n)} \Leftrightarrow \textbf{A}_{(n^c;n)} = \textbf{D}_{(n^c;n)}\textbf{U}^{(n)\mathsf{T}},
\end{equation}
which scales to several products across an arbitrary number of modes as
\begin{equation}\label{CCOT_eq:2.6}
    \textbf{A}_{(n^c;n)} = \left( \bigotimes_{\substack{k = 1 \\ k \neq n}} \textbf{U}^{(k)}\right) \textbf{D}_{(n^c;n)} \textbf{U}^{(n)\mathsf{T}} ,
\end{equation}
for $n \in \{1,\dots,d\}$.

Considering the ML rank-($L_1,\dots,L_{\hat{d}},\cdot,\dots,\cdot$) term decomposition (i.e. BTD) for some tensor $\boldsymbol{\mathcal{A}} \in \mathbb{F}^{I_1 \times \cdots \times I_d}$,
\begin{equation} \label{CCOT_eq:2.9}
    \boldsymbol{\mathcal{A}} = \sum_{r=1}^R \boldsymbol{\mathcal{D}}_r \times_1 \textbf{U}^{(1)}_r \cdots \times_{\hat{d}} \textbf{U}^{(\hat{d})}_r, 
\end{equation}
where $\boldsymbol{\mathcal{D}}_r \in \mathbb{F}^{L_1 \times \cdots \times L_{\hat{d}} \times I_{\hat{d}+1} \times \cdots \times I_d}$ and $\textbf{U}_r^{(n)} \in \mathbb{F}^{I_n \times L_{nr}}$ for $n \in \{1,\dots,\hat{d}\}$ and $r \in \{1,\dots,R\}$.  
By Eq. \ref{CCOT_eq:2.6}, the matricized version is given as:
\begin{equation}\label{CCOT_eq:2.12}
    \textbf{A}_{(n^c;n)} = \sum_{r=1}^R \left( \bigotimes_{\substack{l=1 \\ l \neq n}} \textbf{U}_r^{l} \right) \textbf{D}_{r(n^c;n)} \textbf{U}_r^{(n)\mathsf{T}} 
\end{equation}
The matricized version can be further simplified from the summation form by defining $\textbf{U}^{(n)}$ to be the concatenated factor matrices, $\textbf{U}_r^{(n)}$, of $\boldsymbol{\mathcal{A}}$.
\begin{equation}\label{CCOT_eq:2.10}
    \textbf{U}^{(n)} := [\textbf{U}_1^{(n)} \dots \textbf{U}_R^{(n)}] \in \mathbb{F}^{I_n \times \sum_{r=1}^R L_{nr}}, 
\end{equation}
for $n \in \{1,\dots,\hat{d}\}$, where $\hat{d} \neq d$, $\textbf{U}^{(n)}$ includes
\begin{equation}
    \textbf{U}^{(n)} := [\textbf{I}_{I_n} \dots \textbf{I}_{I_n}] \in \mathbb{F}^{I_n \times RI_n},      
\end{equation}
in the concatenation of the factor matrices for $n \in \{ \hat{d}+1,\dots, d \}$.  

By collecting the mode-$n$ matricizations of the core tensors $\boldsymbol{\mathcal{D}}_r$ in a block-diagonal matrix $\text{Bdiag}(\textbf{D}_{1(n^c;n)}, \dots, \textbf{D}_{R(n^c;c)})$
and by defining the series of Kronecker products as
\begin{equation}\label{CCOT_eq:2.13}
    \bigodot_{\substack{l=1 \\ l \neq n}}^d \textbf{U}^{(l)} := \left[ \bigotimes_{\substack{l=1 \\ l \neq n}}^d \textbf{U}_1^{(l)} \cdots \bigotimes_{\substack{l=1 \\ l \neq n}}^d \textbf{U}_R^{(l)} \right],
\end{equation}
Eq. \ref{CCOT_eq:2.12} can be written as
\begin{equation}\label{CCOT_eq:2.12a}
        \textbf{A}_{(n^c;n)} = \left( \bigodot_{\substack{l=1 \\ l \neq n}}^d \textbf{U}^{(l)} \right) \text{Bdiag}(\textbf{D}_{1(n^c;n)}, \dots, \textbf{D}_{R(n^c;c)}) \textbf{U}^{(n)\mathsf{T}}, 
\end{equation}
for $n \in \{1,\dots,d\}$.

\subsubsection{Idealized problem setting}
Consider some tensor $\boldsymbol{\mathcal{B}} \in \mathbb{F}^{I_1 \times \cdots \times I_d}$ that can be factorized into the same ML rank-($L_{1r}, \dots, L_{\hat{d}r}, \cdot, \dots, \cdot$), possibly scaled by some $\lambda$-term, as some tensor $\boldsymbol{\mathcal{A}}$, i.e.
\begin{equation}\label{CCOT_eq:2.14}
    \boldsymbol{\mathcal{B}} = \sum_{r=1}^R \lambda_r \boldsymbol{\mathcal{D}}_r \bullet_1 \textbf{U}_r^{(1)} \cdots \bullet_{\hat{d}} \textbf{U}_r^{(\hat{d})}, \quad \lambda_1, \dots, \lambda_R \neq 0
\end{equation}
Then by Eq. \ref{CCOT_eq:2.12a}

\begin{equation}
    \scalebox{0.6}{$
    \begin{aligned}
        \textbf{B}_{(n^c;n)} 
        &= \left( \bigodot_{\substack{l=1 \\ l \neq n}}^d \textbf{U}^{(k)} \right) 
        \text{Bdiag}(\lambda_1 \textbf{D}_{1(n^c;n)}, \dots, \lambda_R \textbf{D}_{R(n^c;c)}) 
        \textbf{U}^{(n)T} \\
        &= \left( \bigodot_{\substack{l=1 \\ l \neq n}}^d \textbf{U}^{(k)} \right) \text{Bdiag}(\textbf{D}_{1(n^c;n)}, \dots, \textbf{D}_{R(n^c;c)}) 
        \text{Bdiag} (\lambda_1 \textbf{I}_{L_{n1}}, \dots, \lambda_R \textbf{I}_{L_{nR}}) \textbf{U}^{(n)T}.
    \end{aligned}
    $}
\end{equation}

By assuming the factor matrices, $\textbf{U}^1,\dots, \textbf{U}^{\hat{d}}$ for $\hat{d} \geq 2$, have full column rank, it is easy to show that the first $\hat{d}$ matrix representations of $\boldsymbol{\mathcal{A}}$ and $\boldsymbol{\mathcal{B}}$ have coinciding column spaces:
\begin{equation}\label{CCOT_eq:2.17}
    \text{col}(\textbf{A}_{(n^c;n)}) = \text{col}(\textbf{B}_{(n^c;n)}), \quad n \in \{1,\dots,\hat{d}\}
\end{equation}
Limiting the assumption further to the case where the factor matrices are square and nonsingular,
it follows that,
\begin{equation}\label{CCOT_eq:2.19}
    \textbf{B}_{(n^c;n)} = \textbf{A}_{(n^c;n)} \textbf{M}_n,
\end{equation}
where,
\begin{equation}\label{CCOT_eq:2.20}
    \textbf{M}_n = (\textbf{U}^{(n)\mathsf{T}})^{-1} \text{Bdiag}(\lambda_1 \textbf{I}_{L_{n1}},\dots, \lambda_R \textbf{I}_{L_{nR}}) \textbf{U}^{(n)\mathsf{T}},
\end{equation}
for $n \in \{1,\dots,\hat{d})\}$.

Thus, if all assumptions hold, the column spaces of the first $\hat{d}$ matrix representations of $\boldsymbol{\mathcal{A}}$ and $\boldsymbol{\mathcal{B}}$ coincide, the matrices $\textbf{M}_n := \textbf{A}_{(n^c;n)}^{\dag} \textbf{B}_{(n^c;n)}$ can be diagonalized and have the same spectrum $\lambda_1, \dots, \lambda_R \in \mathbb{F}$ for $n = 1,\dots,\hat{d}$.  
In this arrangement, the concatenated factor matrices $\textbf{U}^{(n)}$ and the sizes of blocks $L_{nr}$, i.e. the overall decompositions of $\boldsymbol{\mathcal{A}}$ and $\boldsymbol{\mathcal{B}}$, can be recovered from the EVDs of $\textbf{M}_1, \dots, \textbf{M}_{\hat{d}}$.
This observation can be leveraged for structural DC as it provides a method for determining the number of terms and associated multilinear ranks of each term in a BTD implicitly using stable methods from numerical linear algebra.
See \cite[Section 4]{domanovComputationComparisonTensor2021} for details and proofs.

\subsubsection{Application: Blind Source Separation Facilitated by Tensorization.}

Determining whether two observed signals share underlying components is a basic problem in signal processing.  
This problem is often difficult due to lack of prior knowledge regarding the component signals and varying amplitudes of component signals between observed signals.  
Domanov and De Lathauwer use the problem of underdetermined blind source separation (BSS) as illustrative application of their proposed method where the goal is to recover source signals given a linear mixture \cite{comonHandbookBlindSource2010}. 
The linear BSS problem consists of the decomposition of an observed data matrix $\textbf{X} \in \mathbb{K}^{R\times N}$ as
\begin{equation} \label{eq:BSS}
    \textbf{X} = \textbf{MS} 
\end{equation}
where $\textbf{M} \in \mathbb{K}^{K \times R}$ is the mixing matrix and $\textbf{S} \in \mathbb{K}^{R \times N}$ is the observed source matrix.  
For each signal there are $N$ samples available.    
A matrix $\textbf{N} \in \mathbb{K}^{R \times N}$ can be incorporated to represent additive noise.  

Underdetermined BSS problems are generally difficult to solve and occur when there are a large number of source signals that are sparsely combined in the observed mixed signals, i.e. a mixing matrix with more columns than rows.  Conversely, overdetermined BSS result in square or skinny mixing matrices and an easier problem to solve.  
One approach to solving an underdetermined BSS problem is to recast the problem as a set of smaller overdetermined BSS problems.
This is accomplished by breaking the problem into a multilabel classification problem where mixtures $i$ and $j$ are placed in the same class if mixture $i$ is generated by (some of) the sources that appear in mixture $j$.

The authors rely on \cite[Theorem 4.1]{domanovComputationComparisonTensor2021} and the assumption that the source signals are reasonably modeled as exponential functions, which allows the sources to be mapped into low ML rank tensors via a process known as tensorization \cite{debalsStochasticDeterministicTensorization2015}. 
Tensorization provides a means to impose certain assumptions about the data as well as to access tensor methods otherwise inapplicable in matrix based formulations such as BSS problem.
The deliberate transformation of data into tensor form is another example of the malleability of tensor problem formulation where the data is tailored to the method (as opposed to the GCP that tailors the method to the data).  
There are a multitude of tensorizations specific to particular families of functions including \emph{L\"ownerization} for rational functions and \emph{Segmentation} for the family of periodic signals \cite{debalsTensorizationApplicationsBlind2017}. 
Domonav and De Lathauwer leverage their assumption in regards to source signals to apply \emph{Hankelization}, a Hankel matrix based tensorization specific to the family of exponential functions.

A Hankel matrix $\textbf{H}$ for an exponential signal $f(k) = az^k$ is a rank-1 matrix given as 
\begin{equation} \label{eq:Hankel Matrix}
    \scalebox{0.88}{$
    \textbf{H} = 
    \begin{bmatrix}
        f(0) & f(1) & f(2) & \cdots \\
        f(1) & f(2) & f(3) & \cdots \\
        f(2) & f(3) & f(4) & \cdots \\
        \vdots & \vdots & \vdots 
    \end{bmatrix} =
    a 
    \begin{bmatrix}
        1 \\ z \\ z^2 \\ \vdots
    \end{bmatrix}
    \begin{bmatrix}
        1 & z & z^2 & \cdots
    \end{bmatrix}.
    $}
\end{equation}

The family of exponential polynomials generalize such simple exponential functions as any combination of sums or products of exponential, sinusoids, and polynomials.
The general form of a univariate exponential polynomial is:
\begin{equation} \label{CCOT_eq:5.1}
    s(t) = \sum_{f=1}^F p_f(t)a^t_f,
\end{equation}
where $p_1,\dots,p_F$ are non-zero polynomials in a single variable and $a_1,\dots,a_F \in \mathbb{C} \setminus \{0\}$.
The Hankel matrix based tensorization technique takes advantage of the fact that the Hankel matrix for an exponential polynomial signal of degree $\delta$ will be of rank $\delta$ \cite{vandevoordeFastExponentialDecomposition1998}. 

Hankelization maps each row of the observed data in $\textbf{X}$ to a Hankel matrix that is stacked in a third order tensor, $\boldsymbol{\mathcal{H}}_{\textbf{x}}$, given as
\begin{equation}
    \boldsymbol{\mathcal{H}}_{\textbf{X}} = \sum_{r=1}^{R} \textbf{H}_{s_r} \otimes \textbf{m}_r = \sum_{r=1}^{R} (\textbf{A}_r \textbf{B}_r^{\mathsf{T}}) \otimes \textbf{m}_r
\end{equation}
where $\textbf{H}_{s_r}$ is the Hankel matrix associated with the $r$-th source, $\textbf{s}_r$. 
Through the lens of linearity, the latter transition can be considered as a hypothesis that the $r$-th source can be approximated by some exponential polynomial of degree $L_{s_r}$ (presumably low) which implies each $\textbf{H}_{s_r}$ has rank $L_{s_r}$ which can be expressed as a product of full column-rank matrices, $\textbf{A} \in \mathbb{R}^{I \times L_{s_r}}$ and $\textbf{B} \in \mathbb{R}^{J \times L_{s_r}}$.  
\begin{figure}[t]
    \centering
    \includegraphics[height=2in]{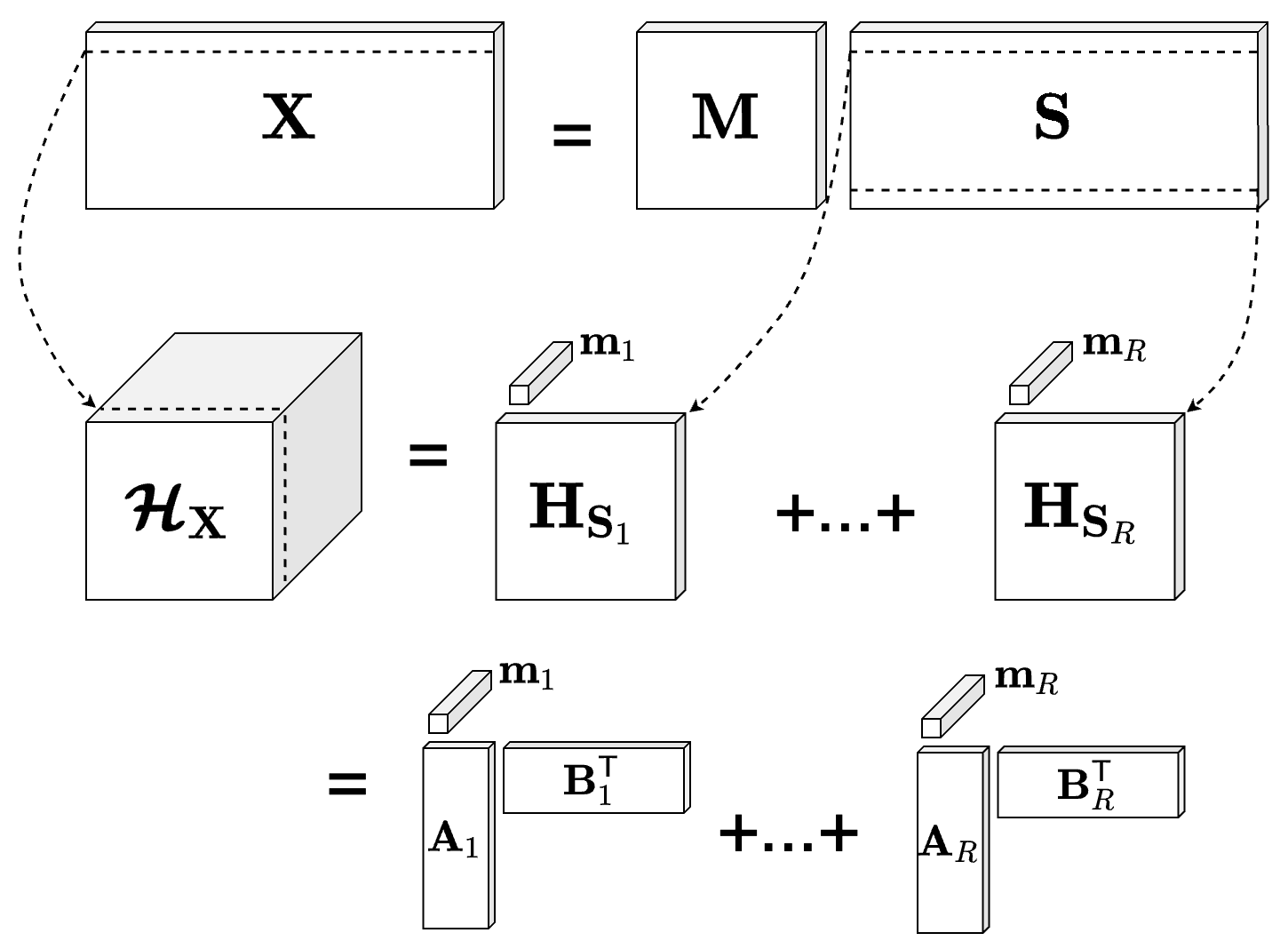}
    \caption{Hankelization of the BSS problem (top and middle) and rank-($L_s, L_s, 1$) block term decomposition (bottom).}
    \label{CCOT_fig:Tensorization}
\end{figure}
The process is mildly bounded by the term $L_s$ that connects the multilinear rank of resulting tensor.

Let $T_s$ denote the sampling time and let $N$ be the number of sampling points,  \cite{debalsTensorizationApplicationsBlind2017,delathauwerBlindSeparationExponential2011} show that for any positive integers $I_1,I_2,I_3$ that sum to $N+2$ and are greater than or equal to $L_s$, the vector $\textbf{s} = [s_1 \dots s_N]^{\mathsf{T}} := [s(0) \dots s((N-1)T_s)]^{\mathsf{T}}$ can be mapped to an $I_1 \times I_2 \times I_3$ ML rank-($L_s,L_s,L_s$) tensor $\boldsymbol{\mathcal{S}}$, where,
\begin{equation} \label{CCOT_eq:5.2}
    L_s := F + \sum_{f=1}^F \deg p_f,
\end{equation}
does not depend on $I_1,I_2,I_3$.

The mapping $H : \textbf{s} \mapsto \boldsymbol{\mathcal{S}}, H : \mathbb{C}^N \rightarrow \mathbb{C}^{I_1 \times I_2 \times I_3}$ is defined as 
\begin{equation} 
    (\boldsymbol{\mathcal{S}})_{i_1 i_2 i_3} = s_{i_1 + i_2 + i_3 - 2} = s((i_1 + i_2 + i_3 - 3) T_s),
\end{equation}
where $1 \leq i_1 \leq I_1, 1 \leq i_2 \leq I_2, 1 \leq i_3 \leq I_3$. \cite{debalsTensorizationApplicationsBlind2017} refers to the mapping $H$ as a "Hankelization" since $(\boldsymbol{\mathcal{S}})_{i_1 i_2 i_3}$ is only dependent on $i_1 + i_2 + i_3$. 

Thus given the linear mapping $H$, if $\textbf{y} = [ y_1 \dots y_N ]^{\mathsf{T}} := [y(0) \dots y((N-1)T_s)]^{\mathsf{T}}$ is a linear mixture of sampled sources with the form given by Eq. \ref{CCOT_eq:5.1}
\begin{equation} \label{CCOT_eq:5.3}
    y(t) =  \textbf{g}_1 s_1(t) + \dots + \textbf{g}_R s_r(t), 
\end{equation}
for $t = 0, T_s, \dots, (N-1)T_s$ and $\min (I_1, I_2, I_3) \geq \max L_{s_r}$, then by Eq. \ref{CCOT_eq:5.2} the decomposition of tensor $\boldsymbol{\mathcal{Y}}$ into a sum of ML rank-$(L_{s_r},L_{s_r},L_{s_r})$ terms is given as
\begin{equation} \label{CCOT_eq:5.4}
    \begin{split}
    \boldsymbol{\mathcal{Y}} := H(\textbf{y}) &= \textbf{g}_1 H(\textbf{s}_1) + \dots + \textbf{g}_R H(\textbf{s}_R) \\
    &= \textbf{g}_1 \boldsymbol{\mathcal{S}}_1 + \dots + \textbf{g}_R \boldsymbol{\mathcal{S}}_R.
    \end{split}
\end{equation}

To illustrate the proposed approach, \cite{domanovComputationComparisonTensor2021} generate 25 mixtures
\begin{equation} \label{CCOT_eq:5.5}
    y_j(t) = \textbf{g}_{1j}s_1(t) + \dots + \textbf{g}_{8j}s_8(t), \quad j = 1,\dots,25.
\end{equation}
using the eight exponential polynomials.

\begin{figure}[t]
    \centering
    \includegraphics[width=.9\columnwidth]{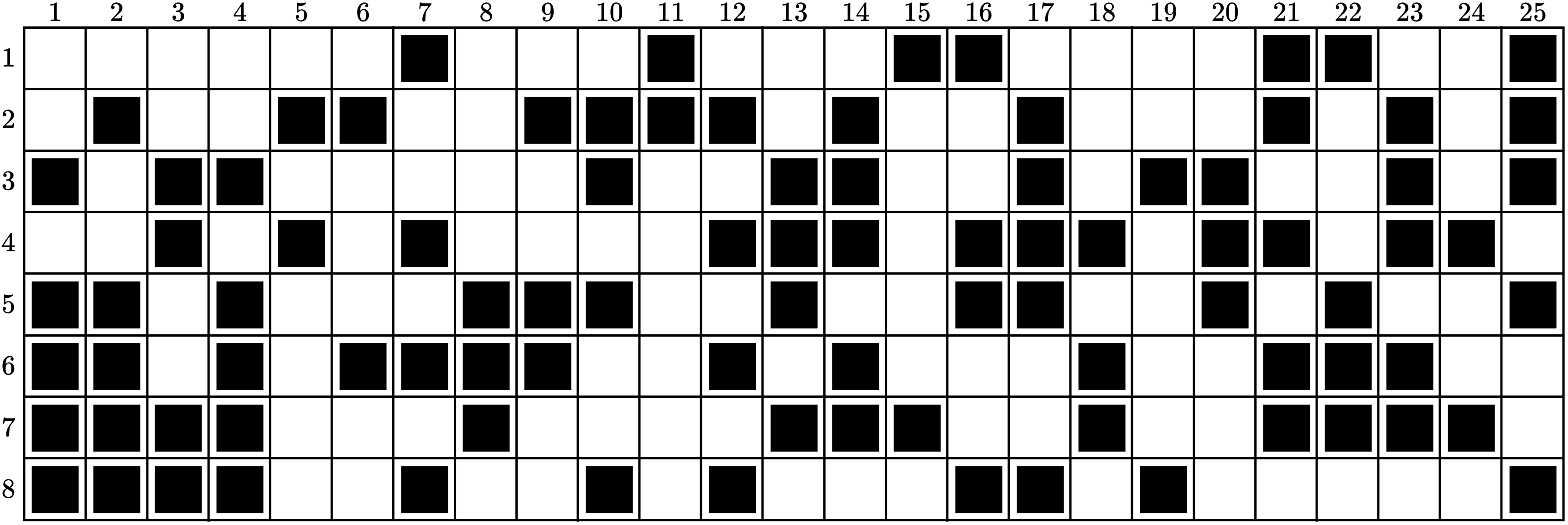}
    \caption{Non-zero pattern of the matrix $\textbf{G}$. Black squares indicate non-zero values in the mixing matrix $\textbf{G}$ }
    \label{CCOT_fig:5.1}
\end{figure}

Coefficients for each of the polynomial terms, $\textbf{g}_{ij}$, are randomly generated such that at least three and at most six of $\textbf{g}_{1j}, \dots, \textbf{g}_{8j}$ are non-zero values drawn from $[-2.5,-0.5] \cup [0.5, 2.5]$ which yields,
\begin{equation} \label{CCOT_eq:5.6}
    [y_1(t) \dots y_{25}(t)] = [s_1(t) \dots s_8(t)]\textbf{G},
\end{equation}
where $\textbf{G} = (\textbf{g}_{ij})$ is an $8\times25$ sparse matrix. 
\ref{CCOT_fig:5.1} shows the non-zero pattern of $\textbf{G}$.
The results presented in \cite{domanovComputationComparisonTensor2021} considers a noisy sampled version of Eq. \ref{CCOT_eq:5.6} with $T_s = 0.05$ and $N=100$:
\begin{equation} \label{CCOT_eq:5.11}
    [\textbf{y}_1^n \dots \textbf{y}_{25}^n] := [\textbf{y}_1 \dots \textbf{y}_{25}] + \sigma\textbf{N} = [\textbf{s}_1 \dots \textbf{s}_8]\textbf{G} + \sigma\textbf{N},
\end{equation}
where the entries of the matrix $\textbf{N}$ are independently drawn from the standard normal distribution $N(0,1)$ and $\sigma = \frac{\| \textbf{y}_1 \dots \textbf{y}_{25} \|_F}{\|\textbf{N}\|_F}$.  
\cite[Theorem 4.1]{domanovComputationComparisonTensor2021} is applied to verify whether any pair of mixtures $(y_i(t),y_j(t))$ is generated by a shared subset of sources, $1 \leq i,j \leq 25, i \neq j$.  

The mixtures $y_1(t), \dots, y_{25}(t)$ are associated with vertices of a directed graph to facilitate visualization where a directed edge from vertex $i$ to vertex $j$ indicates that $y_i(t)$ is generated by sources that are also present in $y_j(t)$.
The goal of this experiment is to recover the nonzero pattern of the mixing matrix, Figure \ref{CCOT_fig:5.1}, without estimating $\textbf{s}_1,\dots,\textbf{s}_8$.  
The approximate decomposition of $\boldsymbol{\mathcal{Y}}_i^n$ into a sum of ML rank-$(L_{s_r},L_{s_r},L_{s_r})$ terms is achieved by defining $H : \mathbb{C}^{100} \rightarrow \mathbb{C}^{34 \times 34 \times 34}$ as the Hankelization mapping, Eq. \ref{CCOT_eq:5.4}, and Eq. \ref{CCOT_eq:5.11}.  
It is given as
\begin{equation}
    \begin{split}
    \boldsymbol{\mathcal{Y}}_i^n :&= H(\textbf{y}_i^n) = \textbf{g}_{i1}H(\textbf{s}_1) + \dots + \textbf{g}_{i8}H(\textbf{s}_8) + \sigma H(\textbf{n}_i) \\ 
    &= \textbf{g}_{i1}\boldsymbol{\mathcal{S}}_1 + \dots + \textbf{g}_{i8}\boldsymbol{\mathcal{S}}_R + \sigma H(\textbf{n}_i)
    \end{split}
\end{equation}

It is trivial to verify the exact values of $L_{s_1},\dots,L_{s_8}$ are 1, 2, 2, 2, 4, 4, 4, 4 respective to said order. For example, applying Euler's formula to $s_8(t)$ leads to

\begin{equation}
    \begin{split}
    s_8(t) &= t\frac{1}{2}\left(e^{(14\pi t - 0.5)i} + e^{-(14\pi t - 0.5)i}\right) \\
    &= (\frac{1}{2} e^{-0.5i} t^1)e^{14 \pi t} + (\frac{1}{2} e^{0.5i} t^1)e^{-14 \pi t},
    \end{split}
\end{equation}

a function that is the sum of 2 deg-1 exponential polynomials, e.g. $L_{s_8} = 2 + (1 + 1) = 4$.  
The authors note that the ML rank-$(4,4,4)$ tensors $\boldsymbol{\mathcal{S}}_5, \dots, \boldsymbol{\mathcal{S}}_8$ can be approximated by ML rank-$(2,2,2)$ tensors with a relative error less than 0.061, which is below the noise level.
The problem of showing that $\boldsymbol{\mathcal{Y}}_i^n$ is generated by the ML rank terms present in $\boldsymbol{\mathcal{Y}}_j^n$ in order to sufficiently verify the source signals present in $y_i(t)$ are also present in $y_j(t)$ is reduced via \cite[Theorem 4.1]{domanovComputationComparisonTensor2021} to verifying 
$$
    \text{col}(\textbf{Y}^n_{i(2,3;1)}) \subseteq \text{col}(\textbf{Y}^n_{j(2,3;1)})
$$

Comparison is accomplished using the first $r_i$ singular vectors $\textbf{u}_{i1},\dots,\textbf{u}_{ir_i}$ of $\textbf{Y}^n_{i(2,3;1)}$ where $r_i$ is the rank of $\textbf{Y}^n_{i(2,3;1)}$ as estimated by a ratio of the $k$th and $(k+1)$st singular values of $\textbf{Y}^n_{i(2,3;1)}$ and a threshold parameter, $\tau$ ( $\tau = 2.3$ was used in the experiment).
This information was used to generate a directed graph where for all $i,j = 1, \dots, 25$ and $i \neq j$, a directed edge from vertex $i$ to vertex $j$ is added to the graph if the $(r_j + 1)$st singular value of the matrix $[\textbf{u}_{i1} \dots \textbf{u}_{ir_i} \textbf{u}_{j1} \dots \textbf{u}_{jr_j}]$ is less than 0.1.
The adjacency matrix for the resulting graph corresponds to the non-zero pattern of the matrix $\textbf{G}$.
Using their proposed method, the authors were able to accurately detect all the edges of the graph without addition of superfluous edges.

\section{Conclusion}
\label{sec:Conclusion}
In the realm of computing, data serves as the fundamental raw material from which valuable information is derived. While the distinction between data and information may appear subtle, it is inherently transformative. Consider, for example, the stars in the night sky—not merely as points of light, but as an extensive stream of raw data. Through computational processes, instruments such as the astrolabe extract navigational insights, converting unstructured observations into actionable knowledge. As modern society increasingly relies on data-driven algorithms and decision-making frameworks, advancing DC becomes not only advantageous but essential for extracting meaningful and reliable information.

Tensors and tensor methods provide a sophisticated toolkit for the  data understanding which speaks to their broader utility in modern computing.
Beyond a demonstrable role in groundbreaking hardware and software advancements, tensors and tensor methods are well equipped to improve modern DC efforts which would add to their role as drivers of innovation at the frontier of technology.
For instance, data understanding will undoubtedly play a pivotal role in unraveling the complexities of AI, bridging the gap to true explainability by providing an improved grasp upon a fundamental component of such AI systems, the data driving the models. 
In essence, refining how data is approached and understood is key to unlocking a future rich with untapped possibilities and solutions.

\subsection{RQ1: What properties of tensors and TDA methods are well-suited for the DC problem space?}
Tensors provide many points of entry for the exploration and deeper understanding of data through their natural representation of structured data and myriad methods.  
In our study of TDA as it relates to DC, we have identified four primary properties of tensors and tensor methods that capture TDA's potential for improving DC:
\begin{itemize}
    \item Tensors are equipped with a variety of \emph{measures} as virtue of their mathematical definition, e.g. rank \cite{comonGenericTypicalRanks2009}, norms \cite{defantTensorNormsOperator1992}. The property of inherent measure in of itself make tensors highly desirable in terms of DC and can be leveraged in the derivation of a variety of dataset characteristics.
    \item Tensors are \emph{multilinear maps} \cite{limTensorsComputations2021} capable of preserving latent relationships formed by higher-dimension datasets, a property not shared by matrix/vector-based DC methods.
    In a general sense, the purpose of linear algebra is to classify linear operators up to isomorphism and study the simplest representation.
    Linear maps are a special case of linear operators that map from a vector space $V$ to an arbitrary field $\mathbb{K}$ which form the dual space $V^\star$ (i.e. the vector space of covectors).
    Multilinear maps are a generalization of linear maps that allow for coordinate-free represenetation and manifest the transformation rules that constitute changes of basis that facilitate informative exploration of the structures found in latent spaces.
    \item Tensors possess the potential for \emph{malleable} problem formulations that provide an ability to adapt methods to the specific details of datasets of interest.
    The GCP \cite{hongGeneralizedCanonicalPolyadic2020} exemplifies this property as it relates to modifying the tool of tensor decompostion, the example for tensor comparison presented by Domanov and De Lathauwer \cite{domanovComputationComparisonTensor2021} exemplifies malleability in the other direction where tensorization is used to modify the dataset to better accommodate the tool.
    \item TDA methods allow for \emph{implicit formulations} that allow a means to navigate computational challenges associated with TDA (e.g. NP-hardness) while still benefiting from the improved theoretical guarantees like unique factorizations of a dataset.  
    Domanov and De Lathauwer \cite{domanovComputationComparisonTensor2021} and works invloving tensor moments of mixture models \cite{pereiraTensorMomentsGaussian2022, zhangMomentEstimationNonparametric2022} exemplify this by leveraging implicit formulation of tensor decomposition to circumvent ill-conditioning and NP-hardness associated with rank determination.    
\end{itemize}

\subsection{RQ2: To what extent has TDA been explored in the problem space of DC in existing studies?}
The review of existing surverys in Subsection \ref{sec:Tensors:ExistingSurveys} demonstrates the adaptation of tensors and TDA in a wide variety of applications as embodied by advancements and hardware design.
The ubiquity of tensors and TDA in such a wide breadth of problem spaces is highlights how various properties of tensors and mulitilinear algebra can be interwoven to arrive at improved results.
Nonetheless, the number of works targeting the use of tensor for characterizing datasets is limited.
Subsection \ref{sec:Tensors:Tensor-based Characterization} provides a selection of recent works that align with types of dataset characteristics used by the dataset comparison framework covered in Subsection \ref{sec:Dataset Characterization:ComparativeAnalysis}: statistical, structural, and model-based.

The GCP \cite{hongGeneralizedCanonicalPolyadic2020}, the theory for tensor moments of mixture models \cite{pereiraTensorMomentsGaussian2022, zhangMomentEstimationNonparametric2022}, and the tensor ring mixture model \cite{wuTERMModelTensor2023} provide tensor-based methods for the statistical characterization of data through parametric and nonparametric probability density estimation.

Multiple methods with potential for the structural characterization of data were presented in Subsection \ref{sec:Tensors:Structural-based Characterization}.
A variety of methods for tensor clustering provides another (larger toolbox) for the structural characterization of data including co-clustering \cite{biernackiSurveyModelBasedCoClustering2023, chiProvableConvexCoclustering2020, wuGeneralTensorSpectral2016}, multiview clustering \cite{caiJointlyModelingClustering2021, chenRepresentationLearningMultiview2022}, spectral clustering \cite{bensonTensorSpectralClustering2015,  wuGeneralTensorSpectral2016}, and subspace-based clustering \cite{abavisaniDeepMultimodalSubspace2018, zhangLowRankTensorConstrained2015}.
Lastly, an algebraic method for tensor amplification was presented as a means of better identifying the number of components required to sufficiently model a data through low-rank approximation by amplifying the rank structure inherent to a dataset \cite{tokcanAlgebraicMethodsTensor2021}.

The tensor methods with potential application in model-based characterization fall into two categories, those that are akin to landmarking meta-features and those akin to model-structure meta-features.
In the space of landmarking type model-based chacaterization, a variety of well known ML algorithm adapted to handle data tensors were discussed and includes tensor classification models \cite{maruhashiLearningMultiWayRelations2018, phanTensorDecompositionsFeature2010}, support tensor machines \cite{biswasLinearSupportTensor2017, haoLinearSupportHigherOrder2013}, and tensor decision trees \cite{krawczykTensorDecisionTrees2021}.
The incorporation of tensors in NN architectures \cite{kossaifiTNetParametrizingFully2019, newmanStableTensorNeural2018, novikovTensorizingNeuralNetworks2015, shiaoTenGANAdversariallyGenerating2024,  wangTensorNetworksMeet2023} present a yet to be explored option for model-structure type model-based characterization through the analysis of the rank of weights learned in a tenor-based NN, potentially yielding a more insightful characterization of a dataset through its learned model than what is provided by metrics like the depth of a decision tree or number of internal nodes.

These examples illuminate the types of characterizations TDA is capable of measuring as well as the potential for tensors and TDA to provide improved DC.
This capacity is largely due to the properties identified and discussed in the response to RQ1 above.
As previously stated, the success of many TDA method depend on finding a solution to systems of multilinear equations, Eq. \ref{eq:multilinearsystemofeqs}.
It is our hope that this work will motivate further exploration in this problem space.

\subsection{RQ3:  What are the canonical DC methods, e.g. structural, statistical, and model-based methods, and how do TDA methods align with these canonical DC methods?}
TDA methods align and differ from existing techniques for different types of dataset characterization. 
\subsubsection{Alignments}
Subsection \ref{sec:Tensors:Tensor-based Characterization} provides the clearest points of alignment between tensor-based DC and existing techniques come in the form of the use of tensors in statistical characterization (probability density estimation) and structural characterization (subspace analysis, clustering, and rank identification).
Tensor-based clustering is the most well researched process in the DC toolkit.

Techniques for model-based DC have yet to be rigorously pursued, but  existing ML approaches that have been modified to accommodate tensor data as input (e.g. tensor decision trees) or NN architectures that incorporate tensors and tensor methods present points of exploration.
Such approaches have the potential to yield model-based characterizations akin to the landmarking and model-structure meta-features discussed in Subsection \ref{sec:Meta-Features}.

Tensor-based modifications of existing methods is a testament to the tensor capacity for malleable formulation as well as its natural extension of fundamental components in computing (matrices and vectors) as a multi-indexed array.

\subsubsection{Differences}
Differences between tensor-based dataset characterization and existing techniques fall into two categories: beneficial and detrimental.
The beneficial differences arise primarily from the mathematical definition(s) of tensors as a generalization of matrices and vectors and the so-called 'blessing of dimensionality' that accompanies high-dimensional tensor space.
Although tensors incur increased complexity and challenge, they also provide advantages in the context of TDA.
Tensors provide an ability to efficiently preserve and encapsulate the latent spaces and structure formed by multi-relational datasets which are destroyed when higher-order data is artificially associated with $\mathbb{R}^n$ \cite{limTensorsComputations2021}.
This ability can lead to improved analysis and models of the data in many cases.
Additionally, data held in tensor form tends to lead to sparsely distributed.
The combination of sparsity and enhanced structural information makes it easier to distinguish between noise and signal in the data and identify separable patterns.
This combination contributes significantly to the utility and ubiquity of tensor decomposition as versatile tool in TDA for the likes of compression, complex pattern discovery, and data completion.
Other benefits include improved theoretical guarantees such uniqueness under relatively mild conditions for a variety of tensor decompositions \cite{bhaskaraUniquenessTensorDecompositions2013, domanovUniquenessComputationDecomposition2020, sidiropoulosUniquenessMultilinearDecomposition2000}.
This is not the case for most matrix factorizations which may be transformed via \emph{any} nonsingular matrix.
Uniqueness is obviously very desirable in terms of modeling any type characteristics associated with a dataset.

To summarize, the beneficial differences between tensor-based DC techniques and existing approaches include improved data separation, richer feature representations, enhanced data sparsity, complex pattern discovery, improved model performance and compression, increased interpretability, and stronger theoretical guarantees.

Detrimental differences also arise from the theoretical aspects of tensors.  Most notable and nontrivial is that most tensor problems are NP-hard \cite{hillarMostTensorProblems2013}.  
Additionally, the 'curse of dimensionality' (discussed in Subsection \ref{sec:Tensors:Challenges}) gives rise to challenges in the form of computational complexity, scalability constraints, and difficulties with numerical considerations like ill-conditioning.
Implicitly formulated solutions like the illustrative example covered in Subsection \ref{sec:Tensorization} provide a silver lining in terms of circumventing the detrimental aspects of tensor-based DC.

\subsection{RQ4: What contemporary problem spaces could benefit from improved tensor-based DC?}
The prevalence of data-driven technologies in modern ML and AI application yields many problem spaces that would benefit from improvements in DC.
Two major areas of interest are artificial dataset generation and explainability in AI, both of which would rely on a deep and rich understanding of datasets and the many nuances attached.
\begin{itemize}
    \item Artificial Dataset Generation
    \begin{itemize}
        \item incorporation of feed back from a comparative analysis framework like the one discussed in Subsection \ref{sec:Dataset Characterization:ComparativeAnalysis} could provide an improved error signal in the training of a generative model.
        \item improved DC provides insights into answering the algorithmic selection problem (discussed in Subsection \ref{sec:Meta-Features}).
    \end{itemize}
    \item Explainable AI: an ability to more accurately identify and articulate the latent structures formed by datasets has the potential to form a basis for explaining the inner workings of ML and AI applications.
\end{itemize}

\subsection{Closing Thoughts}
The abstraction of tensors and the algebras attached extend the representative power of matrices to ever broadening horizons of innovation and meaningful application.  
This versatility make the tensor an inescapable tool in the computational landscape of the future and equip them to be adequately situated to characterize data in natural ways, across ever expanding bases of application.
The tensor's utility and broad adaptation has lead to considerable efforts at the implementation level to address the associated computational challenges.
As such, tensors are positioned to make a positive impact on the problem of DC and provide new insights and information to practitioners in data-intensive fields.

\section*{Acknowledgments}
The authors thank the support of the NSF (Award number(s): 1849463 \& 2223932), the Laboratory for Physical Sciences, and Boise State University - School of Computing.

\bibliographystyle{IEEEtran}
\bibliography{synth}

\begin{thebibliography}{100}
\providecommand{\url}[1]{#1}
\csname url@samestyle\endcsname
\providecommand{\newblock}{\relax}
\providecommand{\bibinfo}[2]{#2}
\providecommand{\BIBentrySTDinterwordspacing}{\spaceskip=0pt\relax}
\providecommand{\BIBentryALTinterwordstretchfactor}{4}
\providecommand{\BIBentryALTinterwordspacing}{\spaceskip=\fontdimen2\font plus
\BIBentryALTinterwordstretchfactor\fontdimen3\font minus \fontdimen4\font\relax}
\providecommand{\BIBforeignlanguage}[2]{{%
\expandafter\ifx\csname l@#1\endcsname\relax
\typeout{** WARNING: IEEEtran.bst: No hyphenation pattern has been}%
\typeout{** loaded for the language `#1'. Using the pattern for}%
\typeout{** the default language instead.}%
\else
\language=\csname l@#1\endcsname
\fi
#2}}
\providecommand{\BIBdecl}{\relax}
\BIBdecl

\bibitem{jouppiTenLessonsThree2021}
N.~P. Jouppi, D.~Hyun~Yoon, M.~Ashcraft, M.~Gottscho, T.~B. Jablin, G.~Kurian, J.~Laudon, S.~Li, P.~Ma, X.~Ma, T.~Norrie, N.~Patil, S.~Prasad, C.~Young, Z.~Zhou, and D.~Patterson, ``Ten {{Lessons From Three Generations Shaped Google}}'s {{TPUv4i}} : {{Industrial Product}},'' in \emph{2021 {{ACM}}/{{IEEE}} 48th {{Annual International Symposium}} on {{Computer Architecture}} ({{ISCA}})}.\hskip 1em plus 0.5em minus 0.4em\relax Virtual: ACM, 2021-06-14/2021-06-19, pp. 1--14.

\bibitem{seshadriEvaluationEdgeTPU2022}
K.~Seshadri, B.~Akin, J.~Laudon, R.~Narayanaswami, and A.~Yazdanbakhsh, ``An {{Evaluation}} of {{Edge TPU Accelerators}} for {{Convolutional Neural Networks}},'' in \emph{2022 {{IEEE International Symposium}} on {{Workload Characterization}} ({{IISWC}})}.\hskip 1em plus 0.5em minus 0.4em\relax Austin, Texas: IEEE, 2022-11-06/2022-11-08, pp. 79--91.

\bibitem{dzeroskiMultirelationalDataMining2003}
S.~D{\v z}eroski, ``Multi-relational data mining: An introduction,'' \emph{ACM SIGKDD Explorations Newsletter}, vol.~5, no.~1, pp. 1--16, Jul. 2003.

\bibitem{limTensorsComputations2021}
L.-H. Lim, ``Tensors in computations,'' \emph{Acta Numerica}, vol.~30, pp. 555--764, May 2021.

\bibitem{dongarraGuestEditorsIntroduction2000}
J.~Dongarra and F.~Sullivan, ``Guest {{Editors Introduction}} to the {{Top}} 10 {{Algorithms}},'' \emph{Computing in Science \& Engineering}, vol.~2, no.~01, pp. 22--23, 2000.

\bibitem{fawziDiscoveringFasterMatrix2022}
A.~Fawzi, M.~Balog, A.~Huang, T.~Hubert, B.~{Romera-Paredes}, M.~Barekatain, A.~Novikov, F.~J. R.~Ruiz, J.~Schrittwieser, G.~Swirszcz, D.~Silver, D.~Hassabis, and P.~Kohli, ``Discovering faster matrix multiplication algorithms with reinforcement learning,'' \emph{Nature}, vol. 610, no. 7930, pp. 47--53, Oct. 2022.

\bibitem{srivastavaTensaurusVersatileAccelerator2020}
N.~Srivastava, H.~Jin, S.~Smith, H.~Rong, D.~Albonesi, and Z.~Zhang, ``Tensaurus: {{A Versatile Accelerator}} for {{Mixed Sparse-Dense Tensor Computations}},'' in \emph{2020 {{IEEE International Symposium}} on {{High Performance Computer Architecture}} ({{HPCA}})}.\hskip 1em plus 0.5em minus 0.4em\relax San Diego, CA, USA: IEEE, 2020-02-22/2020-02-26, pp. 689--702.

\bibitem{tukeyExploratoryDataAnalysis1977}
J.~W. Tukey, \emph{Exploratory Data Analysis}.\hskip 1em plus 0.5em minus 0.4em\relax Reading, PA: Addison-Wesley, 1977.

\bibitem{riceAlgorithmSelectionProblem1976}
J.~R. Rice, ``The {{Algorithm Selection Problem}}*,'' in \emph{Advances in {{Computers}}}, M.~Rubinoff and M.~C. Yovits, Eds.\hskip 1em plus 0.5em minus 0.4em\relax Elsevier, Jan. 1976, vol.~15, pp. 65--118.

\bibitem{engelsUsingDataMetric1998}
R.~Engels and C.~Theusinger, ``Using a {{Data Metric}} for {{Preprocessing Advice}} for {{Data Mining Applications}},'' \emph{ECAI}, vol.~98, pp. 23--28, 1998.

\bibitem{sohnMetaAnalysisClassification1999}
S.~Y. Sohn, ``Meta analysis of classification algorithms for pattern recognition,'' \emph{IEEE Transactions on Pattern Analysis and Machine Intelligence}, vol.~21, no.~11, pp. 1137--1144, Nov. 1999.

\bibitem{segreraInformationTheoreticMeasuresMetalearning2008}
S.~Segrera, J.~Lucas, and M.~Moreno~Garc{\'i}a, ``Information-{{Theoretic Measures}} for {{Meta-learning}},'' in \emph{Hybrid Artificial Intelligence Systems}.\hskip 1em plus 0.5em minus 0.4em\relax Berlin, Heidelberg: Springer Berlin Heidelberg, Sep. 2008, pp. 458--465.

\bibitem{bensusanCasaBatloPasseig2000}
H.~Bensusan and C.~{Giraud-Carrier}, ``Casa {{Batlo}} is in {{Passeig}} de {{Gracia}} or landmarking the expertise space,'' \emph{Proceedings of the ECML'2000 workshop on Meta-Learning: Building Automatic Advice Strategies for Model Selection and Method Combination}, pp. 29--47, 2000.

\bibitem{bernhardMetalearningLandmarkingVarious2000}
P.~Bernhard, B.~Hilan, and G.-C. Christophe, ``Meta-learning by landmarking various learning algorithms,'' \emph{Proceedings of the Seventeenth International Conference on Machine Learning (ICML'2000)}, pp. 743--750, 2000.

\bibitem{bensusanHigherorderApproachMetalearning2000}
H.~Bensusan, C.~{Giraud-Carrier}, and C.~Kennedy, ``A {{Higher-order Approach}} to {{Meta-learning}} .'' \emph{ILP Work-in-progress reports}, vol.~35, p.~44, Jan. 2000.

\bibitem{pengImprovedDatasetCharacterisation2002}
Y.~Peng, P.~A. Flach, C.~Soares, and P.~Brazdil, ``Improved {{Dataset Characterisation}} for {{Meta-learning}},'' in \emph{Discovery {{Science}}}, ser. Lecture {{Notes}} in {{Computer Science}}, S.~Lange, K.~Satoh, and C.~H. Smith, Eds.\hskip 1em plus 0.5em minus 0.4em\relax Berlin, Heidelberg: Springer, 2002, pp. 141--152.

\bibitem{basuDataComplexityPattern2006}
M.~Basu and T.~K. Ho, \emph{Data Complexity in Pattern Recognition}, ser. Advanced Information and Knowledge Processing.\hskip 1em plus 0.5em minus 0.4em\relax London: Springer, 2006.

\bibitem{smithInstanceLevelAnalysis2014}
M.~R. Smith, T.~Martinez, and C.~{Giraud-Carrier}, ``An instance level analysis of data complexity,'' \emph{Machine Learning}, vol.~95, no.~2, pp. 225--256, May 2014.

\bibitem{songAutomaticRecommendationClassification2012}
Q.~Song, G.~Wang, and C.~Wang, ``Automatic recommendation of classification algorithms based on data set characteristics,'' \emph{Pattern Recognition}, vol.~45, no.~7, pp. 2672--2689, Jul. 2012.

\bibitem{wangImprovedDataCharacterization2015}
G.~Wang, Q.~Song, and X.~Zhu, ``An improved data characterization method and its application in classification algorithm recommendation,'' \emph{Applied Intelligence}, vol.~43, no.~4, pp. 892--912, Dec. 2015.

\bibitem{robnik-sikonjaDatasetComparisonWorkflows2018}
M.~{Robnik-{\v S}ikonja}, ``Dataset comparison workflows,'' \emph{International Journal of Data Science}, vol.~3, no.~2, pp. 126--145, Jan. 2018.

\bibitem{matejkaSameStatsDifferent2017}
J.~Matejka and G.~Fitzmaurice, ``Same {{Stats}}, {{Different Graphs}}: {{Generating Datasets}} with {{Varied Appearance}} and {{Identical Statistics}} through {{Simulated Annealing}},'' in \emph{Proceedings of the 2017 {{CHI Conference}} on {{Human Factors}} in {{Computing Systems}}}.\hskip 1em plus 0.5em minus 0.4em\relax Denver Colorado USA: ACM, 2017-05-06/2017-05-11, pp. 1290--1294.

\bibitem{anscombeGraphsStatisticalAnalysis1973}
F.~J. Anscombe, ``Graphs in {{Statistical Analysis}},'' \emph{The American Statistician}, vol.~27, no.~1, pp. 17--21, 1973.

\bibitem{brysRobustMeasuresTail2006a}
G.~Brys, M.~Hubert, and A.~Struyf, ``Robust measures of tail weight,'' \emph{Computational Statistics \& Data Analysis}, vol.~50, no.~3, pp. 733--759, Feb. 2006.

\bibitem{cucconiNuovoTestNon1968}
O.~Cucconi, ``Un {{Nuovo Test Non Parametrico Per Il Confronto Fra Due Gruppi Di Valori Campionari}},'' \emph{Giornale degli Economisti e Annali di Economia}, vol.~27, no. 3/4, pp. 225--248, March-April, 1968.

\bibitem{lepageCombinationWilcoxonAnsariBradley1971}
{\relax YVES}.~LEPAGE, ``A combination of {{Wilcoxon}}'s and {{Ansari-Bradley}}'s statistics,'' \emph{Biometrika}, vol.~58, no.~1, pp. 213--217, Apr. 1971.

\bibitem{jaskowiakStrategiesBuildingEffective2016}
P.~A. Jaskowiak, D.~Moulavi, A.~C.~S. Furtado, R.~J. G.~B. Campello, A.~Zimek, and J.~Sander, ``On strategies for building effective ensembles of relative clustering validity criteria,'' \emph{Knowledge and Information Systems}, vol.~47, no.~2, pp. 329--354, May 2016.

\bibitem{vinhInformationTheoreticMeasures2010}
N.~X. Vinh, J.~Epps, and J.~Bailey, ``Information {{Theoretic Measures}} for {{Clusterings Comparison}}: {{Variants}}, {{Properties}}, {{Normalization}} and {{Correction}} for {{Chance}},'' \emph{Journal of Machine Learning Research}, vol.~11, no.~95, pp. 2837--2854, 2010.

\bibitem{randObjectiveCriteriaEvaluation1971}
W.~M. Rand, ``Objective {{Criteria}} for the {{Evaluation}} of {{Clustering Methods}},'' \emph{Journal of the American Statistical Association}, vol.~66, no. 336, pp. 846--850, Dec. 1971.

\bibitem{hubertComparingPartitions1985}
L.~Hubert and P.~Arabie, ``Comparing partitions,'' \emph{Journal of Classification}, vol.~2, no.~1, pp. 193--218, Dec. 1985.

\bibitem{fowlkesMethodComparingTwo1983}
E.~B. Fowlkes and C.~L. Mallows, ``A {{Method}} for {{Comparing Two Hierarchical Clusterings}},'' \emph{Journal of the American Statistical Association}, vol.~78, no. 383, pp. 553--569, Sep. 1983.

\bibitem{jaccardDistributionCompareeFlore1901}
P.~Jaccard, ``{Distribution compar{\'e}e de la flore alpine dans quelques r{\'e}gions des Alpes occidentales et orientales.}'' \emph{Bulletin de la Societe Vaudoise des Sciences Naturelles}, vol.~37, pp. 241--272, 1901.

\bibitem{meilaComparingClusteringsInformation2007}
M.~Meil{\u a}, ``Comparing clusterings---an information based distance,'' \emph{Journal of Multivariate Analysis}, vol.~98, no.~5, pp. 873--895, May 2007.

\bibitem{kaufmanFindingGroupsData2005}
L.~Kaufman and P.~J. Rousseeuw, \emph{Finding Groups in Data: An Introduction to Cluster Analysis}, ser. Wiley Series in Probability and Mathematical Statistics.\hskip 1em plus 0.5em minus 0.4em\relax Hoboken, N.J: Wiley, 2005.

\bibitem{gowerGeneralCoefficientSimilarity1971}
J.~C. Gower, ``A {{General Coefficient}} of {{Similarity}} and {{Some}} of {{Its Properties}},'' \emph{Biometrics}, vol.~27, no.~4, pp. 857--871, 1971.

\bibitem{ricciMethodesCalculDifferentiel1900}
M.~M.~G. Ricci and T.~{Levi-Civita}, ``{M{\'e}thodes de calcul diff{\'e}rentiel absolu et leurs applications},'' \emph{Mathematische Annalen}, vol.~54, no.~1, pp. 125--201, Mar. 1900.

\bibitem{koldaTensorDecompositionsApplications2009}
T.~G. Kolda and B.~W. Bader, ``Tensor {{Decompositions}} and {{Applications}},'' \emph{SIAM Review}, vol.~51, no.~3, pp. 455--500, Aug. 2009.

\bibitem{domanovUniquenessComputationDecomposition2020}
I.~Domanov and L.~De~Lathauwer, ``On {{Uniqueness}} and {{Computation}} of the {{Decomposition}} of a {{Tensor}} into {{Multilinear Rank-}}\$(1,{{L}}\_r,{{L}}\_r)\$ {{Terms}},'' \emph{SIAM Journal on Matrix Analysis and Applications}, vol.~41, no.~2, pp. 747--803, Jan. 2020.

\bibitem{kruskalThreewayArraysRank1977}
J.~B. Kruskal, ``Three-way arrays: Rank and uniqueness of trilinear decompositions, with application to arithmetic complexity and statistics,'' \emph{Linear Algebra and its Applications}, vol.~18, no.~2, pp. 95--138, Jan. 1977.

\bibitem{domanovComputationComparisonTensor2021}
I.~Domanov and L.~De~Lathauwer, ``From {{Computation}} to {{Comparison}} of {{Tensor Decompositions}},'' \emph{SIAM Journal on Matrix Analysis and Applications}, vol.~42, no.~2, pp. 449--474, Jan. 2021.

\bibitem{baderAlgorithm862MATLAB2006}
B.~W. Bader and T.~G. Kolda, ``Algorithm 862: {{MATLAB}} tensor classes for fast algorithm prototyping,'' \emph{ACM Transactions on Mathematical Software}, vol.~32, no.~4, pp. 635--653, Dec. 2006.

\bibitem{baderEfficientMATLABComputations2008}
------, ``Efficient {{MATLAB Computations}} with {{Sparse}} and {{Factored Tensors}},'' \emph{SIAM Journal on Scientific Computing}, vol.~30, no.~1, pp. 205--231, Jan. 2008.

\bibitem{changMultiRelationalDataCharacterization2021}
S.~Y. Chang and H.-C. Wu, ``Multi-{{Relational Data Characterization}} by {{Tensors}}: {{Tensor Inversion}},'' \emph{IEEE Transactions on Big Data}, pp. 1--1, 2021.

\bibitem{chenTensorSpectralPnorm2020}
B.~Chen and Z.~Li, ``On the tensor spectral p-norm and its dual norm via partitions,'' \emph{Computational Optimization and Applications}, vol.~75, no.~3, pp. 609--628, Apr. 2020.

\bibitem{friedlandSpectralNormSymmetric2020}
S.~Friedland and L.~Wang, ``Spectral norm of a symmetric tensor and its computation,'' \emph{Mathematics of Computation}, vol.~89, no. 325, pp. 2175--2215, Sep. 2020.

\bibitem{signorettoNuclearNormsTensors2011}
M.~Signoretto, L.~Lathauwer, and J.~Suykens, ``Nuclear {{Norms}} for {{Tensors}} and {{Their Use}} for {{Convex Multilinear Estimation}},'' \emph{undefined}, 2011.

\bibitem{comonGenericTypicalRanks2009}
P.~Comon, J.~M.~F. {ten Berge}, L.~De~Lathauwer, and J.~Castaing, ``Generic and typical ranks of multi-way arrays,'' \emph{Linear Algebra and its Applications}, vol. 430, no.~11, pp. 2997--3007, Jun. 2009.

\bibitem{hastadTensorRankNPcomplete1990}
J.~H{\aa}stad, ``Tensor rank is {{NP-complete}},'' \emph{Journal of Algorithms}, vol.~11, no.~4, pp. 644--654, Dec. 1990.

\bibitem{koldaMultilinearOperatorsHigherorder2006}
T.~G. Kolda, ``Multilinear operators for higher-order decompositions.'' {Sandia National Laboratories (SNL), Albuquerque, NM, and Livermore, CA (United States)}, Tech. Rep. SAND2006-2081, 923081, Apr. 2006.

\bibitem{lathauwerMatrixTensorMultilinear1997}
L.~Lathauwer and B.~De~Moor, ``From matrix to tensor: {{Multilinear}} algebra and signal processing,'' \emph{Mathematics in Signal Processing IV}, pp. 1--15, Jan. 1997.

\bibitem{delathauwerMultilinearSingularValue2000}
L.~De~Lathauwer, B.~De~Moor, and J.~Vandewalle, ``A {{Multilinear Singular Value Decomposition}},'' \emph{SIAM Journal on Matrix Analysis and Applications}, vol.~21, no.~4, pp. 1253--1278, Jan. 2000.

\bibitem{tomasiComparisonAlgorithmsFitting2006}
G.~Tomasi and R.~Bro, ``A comparison of algorithms for fitting the {{PARAFAC}} model,'' \emph{Computational Statistics \& Data Analysis}, vol.~50, no.~7, pp. 1700--1734, Apr. 2006.

\bibitem{delathauwerDecompositionsHigherOrderTensor2008a}
L.~De~Lathauwer, ``Decompositions of a {{Higher-Order Tensor}} in {{Block Terms}}---{{Part II}}: {{Definitions}} and {{Uniqueness}},'' \emph{SIAM Journal on Matrix Analysis and Applications}, vol.~30, no.~3, pp. 1033--1066, Jan. 2008.

\bibitem{bhaskaraUniquenessTensorDecompositions2013}
A.~Bhaskara, M.~Charikar, and A.~Vijayaraghavan, ``Uniqueness of {{Tensor Decompositions}} with {{Applications}} to {{Polynomial Identifiability}},'' \emph{arXiv:1304.8087 [cs, math, stat]}, Apr. 2013.

\bibitem{zhouEfficientNonnegativeTucker2015}
G.~Zhou, A.~Cichocki, Q.~Zhao, and S.~Xie, ``Efficient {{Nonnegative Tucker Decompositions}}: {{Algorithms}} and {{Uniqueness}},'' \emph{IEEE Transactions on Image Processing}, vol.~24, no.~12, pp. 4990--5003, Dec. 2015.

\bibitem{udellWhyAreBig2019}
M.~Udell and A.~Townsend, ``Why {{Are Big Data Matrices Approximately Low Rank}}?'' \emph{SIAM Journal on Mathematics of Data Science}, vol.~1, no.~1, pp. 144--160, Jan. 2019.

\bibitem{udellBigDataLow2019}
M.~Udell, ``Big {{Data}} is {{Low Rank}},'' \emph{SIAG/OPT Views and News}, vol.~27, no.~1, 2019.

\bibitem{hillarMostTensorProblems2013}
C.~J. Hillar and L.-H. Lim, ``Most {{Tensor Problems Are NP-Hard}},'' \emph{Journal of the ACM}, vol.~60, no.~6, pp. 45:1--45:39, Nov. 2013.

\bibitem{acarUnsupervisedMultiwayData2009}
E.~Acar and B.~Yener, ``Unsupervised {{Multiway Data Analysis}}: {{A Literature Survey}},'' \emph{IEEE Transactions on Knowledge and Data Engineering}, vol.~21, no.~1, pp. 6--20, Jan. 2009.

\bibitem{fanaee-tTensorbasedAnomalyDetection2016}
H.~{Fanaee-T} and J.~Gama, ``Tensor-based anomaly detection: {{An}} interdisciplinary survey,'' \emph{Knowledge-Based Systems}, vol.~98, pp. 130--147, Apr. 2016.

\bibitem{songTensorCompletionAlgorithms2019}
Q.~Song, H.~Ge, J.~Caverlee, and X.~Hu, ``Tensor {{Completion Algorithms}} in {{Big Data Analytics}},'' \emph{ACM Transactions on Knowledge Discovery from Data}, vol.~13, no.~1, pp. 1--48, Jan. 2019.

\bibitem{thanhContemporaryComprehensiveSurvey2023}
L.~T. Thanh, K.~{Abed-Meraim}, N.~L. Trung, and A.~Hafiane, ``A {{Contemporary}} and {{Comprehensive Survey}} on {{Streaming Tensor Decomposition}},'' \emph{IEEE Transactions on Knowledge and Data Engineering}, vol.~35, no.~11, pp. 10\,897--10\,921, Nov. 2023.

\bibitem{papalexakisTensorsDataMining2016}
E.~E. Papalexakis, C.~Faloutsos, and N.~D. Sidiropoulos, ``Tensors for {{Data Mining}} and {{Data Fusion}}: {{Models}}, {{Applications}}, and {{Scalable Algorithms}},'' \emph{ACM Transactions on Intelligent Systems and Technology}, vol.~8, no.~2, pp. 1--44, Oct. 2016.

\bibitem{luSurveyMultilinearSubspace2011}
H.~Lu, K.~N. Plataniotis, and A.~N. Venetsanopoulos, ``A survey of multilinear subspace learning for tensor data,'' \emph{Pattern Recognition}, vol.~44, no.~7, pp. 1540--1551, Jul. 2011.

\bibitem{sidiropoulosTensorDecompositionSignal2017}
N.~D. Sidiropoulos, L.~De~Lathauwer, X.~Fu, K.~Huang, E.~E. Papalexakis, and C.~Faloutsos, ``Tensor {{Decomposition}} for {{Signal Processing}} and {{Machine Learning}},'' \emph{IEEE Transactions on Signal Processing}, vol.~65, no.~13, pp. 3551--3582, Jul. 2017.

\bibitem{sunTensorsModernStatistical2021}
W.~W. Sun, B.~Hao, and L.~Li, ``Tensors in {{Modern Statistical Learning}},'' in \emph{Wiley {{StatsRef}}: {{Statistics Reference Online}}}.\hskip 1em plus 0.5em minus 0.4em\relax John Wiley \& Sons, Ltd, 2021, pp. 1--25.

\bibitem{chaoSurveyMultiviewClustering2021}
G.~Chao, S.~Sun, and J.~Bi, ``A {{Survey}} on {{Multiview Clustering}},'' \emph{IEEE Transactions on Artificial Intelligence}, vol.~2, no.~2, pp. 146--168, Apr. 2021.

\bibitem{chenRepresentationLearningMultiview2022}
M.-S. Chen, J.-Q. Lin, X.-L. Li, B.-Y. Liu, C.-D. Wang, D.~Huang, and J.-H. Lai, ``Representation {{Learning}} in {{Multi-view Clustering}}: {{A Literature Review}},'' \emph{Data Science and Engineering}, vol.~7, no.~3, pp. 225--241, Sep. 2022.

\bibitem{panagakisTensorMethodsComputer2021}
Y.~Panagakis, J.~Kossaifi, G.~G. Chrysos, J.~Oldfield, M.~A. Nicolaou, A.~Anandkumar, and S.~Zafeiriou, ``Tensor {{Methods}} in {{Computer Vision}} and {{Deep Learning}},'' \emph{Proceedings of the IEEE}, vol. 109, no.~5, pp. 863--890, May 2021.

\bibitem{wangTensorNetworksMeet2023}
M.~Wang, Y.~Pan, Z.~Xu, X.~Yang, G.~Li, and A.~Cichocki, ``Tensor {{Networks Meet Neural Networks}}: {{A Survey}} and {{Future Perspectives}},'' May 2023.

\bibitem{liuTensorDecompositionModel2023a}
X.~Liu and K.~K. Parhi, ``Tensor {{Decomposition}} for {{Model Reduction}} in {{Neural Networks}}: {{A Review}} [{{Feature}}],'' \emph{IEEE Circuits and Systems Magazine}, vol.~23, no.~2, pp. 8--28, 2023.

\bibitem{mutiSurveyTensorSignal2007}
D.~Muti and S.~Bourennane, ``Survey on tensor signal algebraic filtering,'' \emph{Signal Processing}, vol.~87, no.~2, pp. 237--249, Feb. 2007.

\bibitem{cichockiTensorDecompositionsSignal2015a}
A.~Cichocki, D.~Mandic, A.-H. Phan, C.~Caiafa, G.~Zhou, Q.~Zhao, and L.~De~Lathauwer, ``Tensor {{Decompositions}} for {{Signal Processing Applications From Two-way}} to {{Multiway Component Analysis}},'' \emph{IEEE Signal Processing Magazine}, vol.~32, no.~2, pp. 145--163, Mar. 2015.

\bibitem{congTensorDecompositionEEG2015}
F.~Cong, Q.-H. Lin, L.-D. Kuang, X.-F. Gong, P.~Astikainen, and T.~Ristaniemi, ``Tensor decomposition of {{EEG}} signals: {{A}} brief review,'' \emph{Journal of Neuroscience Methods}, vol. 248, pp. 59--69, Jun. 2015.

\bibitem{qiNumericalMultilinearAlgebra2007}
L.~Qi, W.~Sun, and Y.~Wang, ``Numerical multilinear algebra and its applications,'' \emph{Frontiers of Mathematics in China}, vol.~2, no.~4, pp. 501--526, Oct. 2007.

\bibitem{changSurveySpectralTheory2013}
K.~Chang, L.~Qi, and T.~Zhang, ``A survey on the spectral theory of nonnegative tensors,'' \emph{Numerical Linear Algebra with Applications}, vol.~20, no.~6, pp. 891--912, 2013.

\bibitem{psarrasLandscapeSoftwareTensor2021}
C.~Psarras, L.~Karlsson, J.~Li, and P.~Bientinesi, ``The landscape of software for tensor computations,'' Mar. 2021.

\bibitem{papalexakisLargeScaleTensor2013}
E.~Papalexakis, U.~Kang, C.~Faloutsos, N.~Sidiropoulos, and A.~Harpale, ``Large {{Scale Tensor Decompositions}}: {{Algorithmic Developments}} and {{Applications}},'' \emph{IEEE Data Eng. Bull.}, vol.~36, no.~3, pp. 59--66, 2013.

\bibitem{daveHardwareAccelerationSparse2021}
S.~Dave, R.~Baghdadi, T.~Nowatzki, S.~Avancha, A.~Shrivastava, and B.~Li, ``Hardware {{Acceleration}} of {{Sparse}} and {{Irregular Tensor Computations}} of {{ML Models}}: {{A Survey}} and {{Insights}},'' \emph{Proceedings of the IEEE}, vol. 109, no.~10, pp. 1706--1752, Oct. 2021.

\bibitem{xiaoSurveyAcceleratingParallel2023}
G.~Xiao, C.~Yin, T.~Zhou, X.~Li, Y.~Chen, and K.~Li, ``A {{Survey}} of {{Accelerating Parallel Sparse Linear Algebra}},'' \emph{ACM Computing Surveys}, vol.~56, no.~1, pp. 21:1--21:38, Aug. 2023.

\bibitem{menghaniEfficientDeepLearning2023}
G.~Menghani, ``Efficient {{Deep Learning}}: {{A Survey}} on {{Making Deep Learning Models Smaller}}, {{Faster}}, and {{Better}},'' \emph{ACM Computing Surveys}, vol.~55, no.~12, pp. 259:1--259:37, Mar. 2023.

\bibitem{VoltaWorldMost2017}
``Inside {{Volta}}: {{The World}}'s {{Most Advanced Data Center GPU}},'' May 2017.

\bibitem{kangWhySystolicArchitectures1982}
H.~Kang, ``Why {{Systolic Architectures}}?'' \emph{Computer}, vol.~15, no.~1, pp. 37--46, 1982.

\bibitem{hongGeneralizedCanonicalPolyadic2020}
D.~Hong, T.~G. Kolda, and J.~A. Duersch, ``Generalized {{Canonical Polyadic Tensor Decomposition}},'' \emph{SIAM Review}, vol.~62, no.~1, pp. 133--163, Jan. 2020.

\bibitem{pereiraTensorMomentsGaussian2022}
J.~M. Pereira, J.~Kileel, and T.~G. Kolda, ``Tensor {{Moments}} of {{Gaussian Mixture Models}}: {{Theory}} and {{Applications}},'' Mar. 2022.

\bibitem{zhangMomentEstimationNonparametric2022}
Y.~Zhang and J.~Kileel, ``Moment {{Estimation}} for {{Nonparametric Mixture Models Through Implicit Tensor Decomposition}},'' https://arxiv.org/abs/2210.14386v1, Oct. 2022.

\bibitem{shermanEstimatingHigherOrderMoments2020}
S.~Sherman and T.~G. Kolda, ``Estimating {{Higher-Order Moments Using Symmetric Tensor Decomposition}},'' \emph{SIAM Journal on Matrix Analysis and Applications}, vol.~41, no.~3, pp. 1369--1387, Jan. 2020.

\bibitem{wuTERMModelTensor2023}
R.~Wu, J.~Liu, C.~Zhu, A.-H. Phan, I.~V. Oseledets, and Y.~Liu, ``{{TERM Model}}: {{Tensor Ring Mixture Model}} for {{Density Estimation}},'' Dec. 2023.

\bibitem{zhaoTensorRingDecomposition2016}
Q.~Zhao, G.~Zhou, S.~Xie, L.~Zhang, and A.~Cichocki, ``Tensor {{Ring Decomposition}},'' Jun. 2016.

\bibitem{jegelkaApproximationAlgorithmsTensor2009}
S.~Jegelka, S.~Sra, and A.~Banerjee, ``Approximation {{Algorithms}} for {{Tensor Clustering}},'' in \emph{Algorithmic {{Learning Theory}}}, R.~Gavald{\`a}, G.~Lugosi, T.~Zeugmann, and S.~Zilles, Eds.\hskip 1em plus 0.5em minus 0.4em\relax Berlin, Heidelberg: Springer Berlin Heidelberg, 2009, vol. 5809, pp. 368--383.

\bibitem{fangComprehensiveSurveyMultiView2023a}
U.~Fang, M.~Li, J.~Li, L.~Gao, T.~Jia, and Y.~Zhang, ``A {{Comprehensive Survey}} on {{Multi-View Clustering}},'' \emph{IEEE Transactions on Knowledge and Data Engineering}, vol.~35, no.~12, pp. 12\,350--12\,368, Dec. 2023.

\bibitem{biernackiSurveyModelBasedCoClustering2023}
C.~Biernacki, J.~Jacques, and C.~Keribin, ``A {{Survey}} on {{Model-Based Co-Clustering}}: {{High Dimension}} and {{Estimation Challenges}},'' \emph{Journal of Classification}, vol.~40, no.~2, pp. 332--381, Jul. 2023.

\bibitem{bensonTensorSpectralClustering2015}
A.~R. Benson, D.~F. Gleich, and J.~Leskovec, ``Tensor {{Spectral Clustering}} for {{Partitioning Higher-order Network Structures}},'' in \emph{Proceedings of the 2015 {{SIAM International Conference}} on {{Data Mining}} ({{SDM}})}, ser. Proceedings.\hskip 1em plus 0.5em minus 0.4em\relax {Society for Industrial and Applied Mathematics}, Jun. 2015, pp. 118--126.

\bibitem{jiaMultiViewSpectralClustering2021}
Y.~Jia, H.~Liu, J.~Hou, S.~Kwong, and Q.~Zhang, ``Multi-{{View Spectral Clustering Tailored Tensor Low-Rank Representation}},'' \emph{IEEE Transactions on Circuits and Systems for Video Technology}, vol.~31, no.~12, pp. 4784--4797, Dec. 2021.

\bibitem{wuGeneralTensorSpectral2016}
T.~Wu, A.~R. Benson, and D.~F. Gleich, ``General {{Tensor Spectral Co-clustering}} for {{Higher-Order Data}},'' \emph{Advances in Neural Information Processing Systems}, vol.~29, pp. 2559--2567, 2016.

\bibitem{chiProvableConvexCoclustering2020}
E.~C. Chi, B.~J. Gaines, W.~W. Sun, H.~Zhou, and J.~Yang, ``Provable {{Convex Co-clustering}} of {{Tensors}},'' \emph{Journal of Machine Learning Research}, vol.~21, no. 214, pp. 1--58, 2020.

\bibitem{tibshiraniRegressionShrinkageSelection1996}
R.~Tibshirani, ``Regression {{Shrinkage}} and {{Selection Via}} the {{Lasso}},'' \emph{Journal of the Royal Statistical Society: Series B (Methodological)}, vol.~58, no.~1, pp. 267--288, Jan. 1996.

\bibitem{heDetectingNumberClusters2010}
Z.~He, A.~Cichocki, S.~Xie, and K.~Choi, ``Detecting the {{Number}} of {{Clusters}} in n-{{Way Probabilistic Clustering}},'' \emph{IEEE Transactions on Pattern Analysis and Machine Intelligence}, vol.~32, no.~11, pp. 2006--2021, Nov. 2010.

\bibitem{tokcanAlgebraicMethodsTensor2021}
N.~Tokcan, J.~Gryak, K.~Najarian, and H.~Derksen, ``Algebraic {{Methods}} for {{Tensor Data}},'' \emph{SIAM Journal on Applied Algebra and Geometry}, vol.~5, no.~1, pp. 1--27, Jan. 2021.

\bibitem{papalexakisMoreViewsGraph2013}
E.~E. Papalexakis, L.~Akoglu, and D.~Ience, ``Do more views of a graph help? {{Community}} detection and clustering in multi-graphs,'' in \emph{Proceedings of the 16th {{International Conference}} on {{Information Fusion}}}.\hskip 1em plus 0.5em minus 0.4em\relax Istanbul, Turkey: IEEE, 2013-07-09/2013-07-12, pp. 899--905.

\bibitem{ouvrardAdjacencyTensorRepresentation2018}
X.~Ouvrard, J.-M.~L. Goff, and S.~{Marchand-Maillet}, ``Adjacency and {{Tensor Representation}} in {{General Hypergraphs Part}} 1: E-adjacency {{Tensor Uniformisation Using Homogeneous Polynomials}},'' \emph{arXiv:1712.08189 [cs, math]}, May 2018.

\bibitem{jiSurveyTensorTechniques2019}
Y.~Ji, Q.~Wang, X.~Li, and J.~Liu, ``A {{Survey}} on {{Tensor Techniques}} and {{Applications}} in {{Machine Learning}},'' \emph{IEEE Access}, vol.~7, pp. 162\,950--162\,990, 2019.

\bibitem{biswasLinearSupportTensor2017}
S.~K. Biswas and P.~Milanfar, ``Linear {{Support Tensor Machine With LSK Channels}}: {{Pedestrian Detection}} in {{Thermal Infrared Images}},'' \emph{IEEE Transactions on Image Processing}, vol.~26, no.~9, pp. 4229--4242, Sep. 2017.

\bibitem{haoLinearSupportHigherOrder2013}
Z.~Hao, L.~He, B.~Chen, and X.~Yang, ``A {{Linear Support Higher-Order Tensor Machine}} for {{Classification}},'' \emph{IEEE Transactions on Image Processing}, vol.~22, no.~7, pp. 2911--2920, Jul. 2013.

\bibitem{krawczykTensorDecisionTrees2021}
B.~Krawczyk, ``Tensor decision trees for continual learning from drifting data streams,'' \emph{Machine Learning}, vol. 110, no.~11, pp. 3015--3035, Dec. 2021.

\bibitem{shiaoTenGANAdversariallyGenerating2024}
W.~Shiao, B.~A. Miller, K.~Chan, P.~Yu, T.~{Eliassi-Rad}, and E.~E. Papalexakis, ``{{TenGAN}}: Adversarially generating multiplex tensor graphs,'' \emph{Data Mining and Knowledge Discovery}, vol.~38, no.~1, pp. 1--21, Jan. 2024.

\bibitem{maruhashiLearningMultiWayRelations2018}
K.~Maruhashi, M.~Todoriki, T.~Ohwa, K.~Goto, Y.~Hasegawa, H.~Inakoshi, and H.~Anai, ``Learning {{Multi-Way Relations}} via {{Tensor Decomposition With Neural Networks}},'' in \emph{Thirty-{{Second AAAI Conference}} on {{Artificial Intelligence}}}, Apr. 2018.

\bibitem{phanTensorDecompositionsFeature2010}
A.~H. Phan and A.~Cichocki, ``Tensor decompositions for feature extraction and classification of high dimensional datasets,'' \emph{Nonlinear Theory and Its Applications, IEICE}, vol.~1, no.~1, pp. 37--68, 2010.

\bibitem{kossaifiTensorRegressionNetworks2017}
J.~Kossaifi, Z.~C. Lipton, A.~Khanna, T.~Furlanello, and A.~Anandkumar, ``Tensor {{Regression Networks}},'' \emph{arXiv:1707.08308 [cs]}, Jul. 2017.

\bibitem{maTensorizedTransformerLanguage2019a}
X.~Ma, P.~Zhang, S.~Zhang, N.~Duan, Y.~Hou, M.~Zhou, and D.~Song, ``A {{Tensorized Transformer}} for {{Language Modeling}},'' in \emph{Advances in {{Neural Information Processing Systems}}}, vol.~32.\hskip 1em plus 0.5em minus 0.4em\relax Curran Associates, Inc., 2019.

\bibitem{kossaifiTNetParametrizingFully2019}
J.~Kossaifi, A.~Bulat, G.~Tzimiropoulos, and M.~Pantic, ``T-{{Net}}: {{Parametrizing Fully Convolutional Nets With}} a {{Single High-Order Tensor}},'' in \emph{Proceedings of the {{IEEE}}/{{CVF Conference}} on {{Computer Vision}} and {{Pattern Recognition}}}, 2019, pp. 7822--7831.

\bibitem{newmanStableTensorNeural2018}
E.~Newman, L.~Horesh, H.~Avron, and M.~Kilmer, ``Stable {{Tensor Neural Networks}} for {{Rapid Deep Learning}},'' \emph{arXiv:1811.06569 [cs, math, stat]}, Nov. 2018.

\bibitem{kilmerFactorizationStrategiesThirdorder2011}
M.~E. Kilmer and C.~D. Martin, ``Factorization strategies for third-order tensors,'' \emph{Linear Algebra and its Applications}, vol. 435, no.~3, pp. 641--658, Aug. 2011.

\bibitem{hutterApplicationPerformanceModeling2023}
E.~Hutter and E.~Solomonik, ``Application {{Performance Modeling}} via {{Tensor Completion}},'' in \emph{Proceedings of the {{International Conference}} for {{High Performance Computing}}, {{Networking}}, {{Storage}} and {{Analysis}}}, ser. {{SC}} '23.\hskip 1em plus 0.5em minus 0.4em\relax New York, NY, USA: Association for Computing Machinery, Nov. 2023, pp. 1--14.

\bibitem{chiTensorsSparsityNonnegative2012}
E.~C. Chi and T.~G. Kolda, ``On {{Tensors}}, {{Sparsity}}, and {{Nonnegative Factorizations}},'' \emph{SIAM Journal on Matrix Analysis and Applications}, vol.~33, no.~4, pp. 1272--1299, Jan. 2012.

\bibitem{ranadiveAllOnceCP2021}
T.~M. Ranadive and M.~M. Baskaran, ``An {{All}}--at--{{Once CP Decomposition Method}} for {{Count Tensors}},'' in \emph{2021 {{IEEE High Performance Extreme Computing Conference}} ({{HPEC}})}, Sep. 2021, pp. 1--8.

\bibitem{gordonGeneralized2Linear2Models2002}
G.~J. Gordon, ``Generalized2 {{Linear2}} models,'' in \emph{Proceedings of the 15th {{International Conference}} on {{Neural Information Processing Systems}}}, ser. {{NIPS}}'02.\hskip 1em plus 0.5em minus 0.4em\relax Vancouver, BC, Canada: MIT Press, Jan. 2002, pp. 593--600.

\bibitem{debalsTensorizationApplicationsBlind2017}
O.~Debals, ``Tensorization and {{Applications}} in {{Blind Source Separation}},'' Ph.D. dissertation, KU Leuven, Leuven, Belgium, Aug. 2017.

\bibitem{comonHandbookBlindSource2010}
P.~Comon and C.~Jutten, Eds., \emph{Handbook of Blind Source Separation: Independent Component Analysis and Applications}, 1st~ed.\hskip 1em plus 0.5em minus 0.4em\relax Amsterdam ; Boston: Elsevier, 2010.

\bibitem{debalsStochasticDeterministicTensorization2015}
O.~Debals and L.~De~Lathauwer, ``Stochastic and {{Deterministic Tensorization}} for {{Blind Signal Separation}},'' in \emph{Proceedings of the 12th {{International Conference}} on {{Latent Variable Analysis}} and {{Signal Separation}} - {{Volume}} 9237}, ser. {{LVA}}/{{ICA}} 2015, vol. 9237.\hskip 1em plus 0.5em minus 0.4em\relax Liberec, Czech Republic: Springer-Verlag, Aug. 2015, pp. 3--13.

\bibitem{vandevoordeFastExponentialDecomposition1998}
D.~Vandevoorde, ``A fast exponential decomposition algorithm and its applications to structured matrices,'' Ph.D. dissertation, Rensselaer Polytechnic Institute, USA, 1998.

\bibitem{delathauwerBlindSeparationExponential2011}
L.~De~Lathauwer, ``Blind {{Separation}} of {{Exponential Polynomials}} and the {{Decomposition}} of a {{Tensor}} in {{Rank-}}\$({{L}}\_r,{{L}}\_r,1)\$ {{Terms}},'' \emph{SIAM Journal on Matrix Analysis and Applications}, vol.~32, no.~4, pp. 1451--1474, Oct. 2011.

\bibitem{defantTensorNormsOperator1992}
A.~Defant and K.~Floret, \emph{Tensor {{Norms}} and {{Operator Ideals}}}.\hskip 1em plus 0.5em minus 0.4em\relax Elsevier, Nov. 1992.

\bibitem{caiJointlyModelingClustering2021}
B.~Cai, J.~Zhang, and W.~W. Sun, ``Jointly {{Modeling}} and {{Clustering Tensors}} in {{High Dimensions}},'' Oct. 2021.

\bibitem{abavisaniDeepMultimodalSubspace2018}
M.~Abavisani and V.~M. Patel, ``Deep {{Multimodal Subspace Clustering Networks}},'' \emph{IEEE Journal of Selected Topics in Signal Processing}, vol.~12, no.~6, pp. 1601--1614, Dec. 2018.

\bibitem{zhangLowRankTensorConstrained2015}
C.~Zhang, H.~Fu, S.~Liu, G.~Liu, and X.~Cao, ``Low-{{Rank Tensor Constrained Multiview Subspace Clustering}},'' in \emph{Proceedings of the {{IEEE International Conference}} on {{Computer Vision}}}.\hskip 1em plus 0.5em minus 0.4em\relax Santiago, Chile: IEEE, 2015-12-07/2015-12-13, pp. 1582--1590.

\bibitem{novikovTensorizingNeuralNetworks2015}
A.~Novikov, D.~Podoprikhin, A.~Osokin, and D.~Vetrov, ``Tensorizing {{Neural Networks}},'' \emph{arXiv:1509.06569 [cs]}, Dec. 2015.

\bibitem{sidiropoulosUniquenessMultilinearDecomposition2000}
N.~D. Sidiropoulos and R.~Bro, ``On the uniqueness of multilinear decomposition of {{N-way}} arrays,'' \emph{Journal of Chemometrics}, vol.~14, no.~3, pp. 229--239, 2000.

\end{thebibliography}


 





\vfill

\end{document}